\documentclass[conference]{IEEEtran}
\usepackage{times}
\usepackage[dvipsnames]{xcolor}
\usepackage[utf8]{inputenc}
\usepackage{bm}
\usepackage{siunitx}
\usepackage{amsmath} % assumes amsmath package installed
\usepackage{amssymb}  % assumes amsmath package installed
\usepackage{amsfonts}
\usepackage{dsfont}
\usepackage[T1]{fontenc}
\usepackage{microtype}
\usepackage{graphicx}
\usepackage{multirow}
\usepackage{subcaption} 
\let\labelindent\relax
\usepackage{enumitem} 
\usepackage[numbers]{natbib}
\usepackage{multicol}
\usepackage[bookmarks=true]{hyperref}
\usepackage{booktabs}
\usepackage{xspace}
\usepackage{makecell} 
\usepackage{caption}
\usepackage{lipsum}
\usepackage{mathtools, cuted}
\definecolor{myPurple}{HTML}{d98bff}
\definecolor{myGreen}{HTML}{32d531}
\definecolor{myYellow}{HTML}{ffe400}
\definecolor{myBlue}{HTML}{02cae2}

\newcommand{\our}{\textsc{HugWBC}\xspace}

\begin{document}
\newtheorem{example}{Example}%[section]
\newtheorem{definition}{Definition}%[section]
\newtheorem{lemma}{Lemma}%[section]
\newtheorem{theorem}{Theorem}%[section]
\newtheorem{proposition}{Proposition}%[section]
\newtheorem{corollary}{Corollary}%[proposition]
\newtheorem{assumption}{Assumption}
\newtheorem{observation}{Observation}

\newcommand{\fig}[1]{Fig.~\ref{#1}}
\newcommand{\eq}[1]{Eq.~(\ref{#1})}
\newcommand{\ineq}[1]{Ineq.~(\ref{#1})}
\newcommand{\tb}[1]{Tab.~\ref{#1}}
\newcommand{\se}[1]{Section~\ref{#1}}
\newcommand{\ap}[1]{Appendix~\ref{#1}}
\newcommand{\pa}[1]{Part~\ref{#1}}
\newcommand{\lm}[1]{Lemma~\ref{#1}}
\newcommand{\prop}[1]{Proposition~\ref{#1}}
\newcommand{\alg}[1]{Algo.~\ref{#1}}
\newcommand{\theo}[1]{Theorem~\ref{#1}}
\newcommand{\defi}[1]{Definition~\ref{#1}}
\newcommand{\assum}[1]{Assumption~\ref{#1}}
\newcommand{\observe}[1]{Observation~\ref{#1}}

\newcommand*{\dif}{\mathop{}\!\mathrm{d}}
\newcommand*{\kl}{\mathrm{KL}}
\newcommand{\bbI}{\ensuremath{\mathbb{I}}} % Indicator
\newcommand{\bbE}{\ensuremath{\mathbb{E}}} % 
\newcommand{\bbS}{\ensuremath{\mathbb{S}}} % 
\newcommand{\bbR}{\ensuremath{\mathbb{R}}} % Real Numbers
\newcommand{\caA}{\ensuremath{\mathcal{A}}} % Action
\newcommand{\caS}{\ensuremath{\mathcal{S}}} % State
\newcommand{\caAt}{\ensuremath{\mathcal{\tilde{A}}}} 
\newcommand{\caSt}{\ensuremath{\mathcal{\tilde{S}}}} 
\newcommand{\caN}{\ensuremath{\mathcal{N}}} % Normal Distribution
\newcommand{\caM}{\ensuremath{\mathcal{M}}} % Model
\newcommand{\caMt}{\ensuremath{\mathcal{\tilde{M}}}}
\newcommand{\caD}{\ensuremath{\mathcal{D}}} 
\newcommand{\caG}{\ensuremath{\mathcal{G}}} 
\newcommand{\caL}{\ensuremath{\mathcal{L}}} 
\newcommand{\caT}{\ensuremath{\mathcal{T}}} 
\newcommand{\caO}{\ensuremath{\mathcal{O}}} 
\newcommand{\caTt}{\ensuremath{\mathcal{\tilde{T}}}}
\newcommand{\caB}{\ensuremath{\mathcal{B}}} 
\newcommand{\kld}{\text{D}_{\text{KL}}} 
\newcommand{\jsd}{\text{D}_{\text{JS}}}
\newcommand{\fd}{\text{D}_{\text{f}}} 
\newcommand{\iter}[2]{{#1}^{(#2)}}
\newcommand{\piE}{{\pi_E}}
\newcommand{\hr}{\hat{r}}
\newcommand{\hpi}{\hat{\pi}}

\newcommand{\hytt}[1]{\texttt{\hyphenchar\font=\defaulthyphenchar #1}}

\newcommand{\cmark}{\ding{51}}%
\newcommand{\xmark}{\ding{55}}%
\newcommand{\tc}[1]{\textcolor{red}{#1}}

\let\titleold\title
\renewcommand{\title}[1]{\titleold{#1}\newcommand{\thetitle}{#1}}
\def\maketitlesupplementary
   {
   \newpage
       \twocolumn[
        \centering
        \Large
        % \textbf{\thetitle}\\
        \textbf{Visual Whole-Body Control for Legged Loco-Manipulation}\\
        \vspace{0.5em}Appendix \\
        \vspace{1.0em}
       ] %< twocolumn
   }
\providecommand{\YJ}[1]{{\color{blue}\textbf{YJ:}{#1}}}

\title{A Unified and General Humanoid Whole-Body Controller for Versatile Locomotion}

% You will get a Paper-ID when submitting a pdf file to the conference system
% \author{Author Names Omitted for Anonymous Review. Paper-ID [265]}

\author{\authorblockN{Yufei Xue\textsuperscript{\dag1,2}
\quad Wentao Dong\textsuperscript{\dag1,2}
\quad Minghuan Liu\textsuperscript{\^{}1}
\quad Weinan Zhang\textsuperscript{1} \quad Jiangmiao Pang\textsuperscript{2}}
\authorblockA{
\textsuperscript{1}Shanghai Jiao Tong University \quad \textsuperscript{2}Shanghai AI Lab \\ \textsuperscript{\dag}Equal Contributions\quad \textsuperscript{\^{}}Project Lead \\
\href{https://hugwbc.github.io}{\texttt{https://hugwbc.github.io}}
}
}
%\author{\authorblockN{Michael Shell}
%\authorblockA{School of Electrical and\\Computer Engineering\\
%Georgia Institute of Technology\\
%Atlanta, Georgia 30332--0250\\
%Email: mshell@ece.gatech.edu}
%\and
%\authorblockA{Twentieth Century Fox\\
%\and
%\authorblockN{James Kirk\\ and Montgomery Scott}
%\authorblockA{Starfleet Academy\\
%San Francisco, California 96678-2391\\

% avoiding spaces at the end of the author lines is not a problem with
% conference papers because we don't use \thanks or \IEEEmembership

%Homer Simpson\authorrefmark{2},
%Montgomery Scott\authorrefmark{3} and
%Eldon Tyrell\authorrefmark{4}}
%Telephone: (800) 555--1212, Fax: (888) 555--1212}
%\authorblockA{\authorrefmark{4}Tyrell Inc., 123 Replicant Street, Los Angeles, California 90210--4321}}

% \maketitle
\makeatletter

\twocolumn[{
\renewcommand\twocolumn[1][]{#1}
\maketitle
\begin{center}
    \vspace{-8mm}
    \includegraphics[width=.88\linewidth]{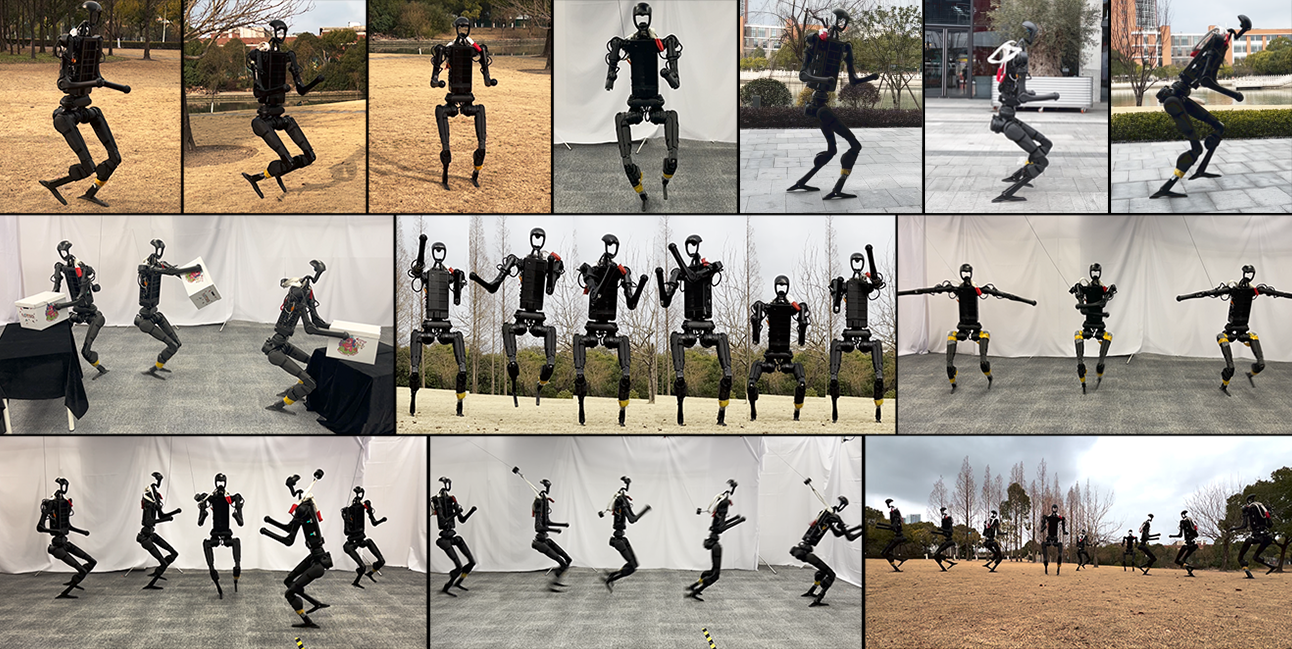}
    \vspace{-6pt}
\end{center}
\captionof{figure}{\small 
\textbf{Humanoid capabilities supported by \our.} \textbf{First row:} \our allows four standard gaits - walking, jumping, standing, and hopping - with multiple customizable parameters to adjust the foot and pose behaviors, using one policy for 3 of the 4 gaits. \textbf{Second row:} \our supports real-time interventions from external upper-body controllers, enabling loco-manipulation while maintaining precise control under any locomotive behavior. \textbf{Third row}: Various command combinations enable the robot to perform highly dynamic movements.
% Combinations of various commands control the robot in high dynamic.
}
% \vspace{-2mm}
\label{fig:teaser}
% \bigbreak
}]
\begin{abstract}
Locomotion is a fundamental skill for humanoid robots. However, most existing works make locomotion a single, tedious, unextendable, and unconstrained movement.
This limits the kinematic capabilities of humanoid robots. In contrast, humans possess versatile athletic abilities--running, jumping, hopping, and finely adjusting gait parameters such as frequency and foot height.
In this paper, we investigate solutions to bring such versatility into humanoid locomotion and thereby propose \our: a unified and general humanoid whole-body controller for versatile locomotion.
By designing a general command space in the aspect of tasks and behaviors, along with advanced techniques like symmetrical loss and intervention training for learning a whole-body humanoid controlling policy in simulation, \our enables real-world humanoid robots to produce various natural gaits, including walking, jumping, standing, and hopping, with customizable parameters such as frequency, foot swing height, further combined with different body height, waist rotation, and body pitch.
Beyond locomotion, \our also supports real-time interventions from external upper-body controllers like teleoperation, enabling loco-manipulation with precision under any locomotive behavior.
Extensive experiments validate the high tracking accuracy and robustness of \our with/without upper-body intervention for all commands, and we further provide an in-depth analysis of how the various commands affect humanoid movement and offer insights into the relationships between these commands.
To our knowledge, \our is the first humanoid whole-body controller that supports such versatile locomotion behaviors with high robustness and flexibility.
\end{abstract}
% Think about us humans, we own versatile athletic abilities, such as running, jumping, and even hopping. Even when walking, we can fine-tune our frequencies, strides, and foot heights.
% However, under the combined action of multiple control signals, directly learning the diverse behavior of humanoid robots can easily lead to unnatural behavior.

\IEEEpeerreviewmaketitle

\section{Introduction}
Recent progress in humanoid robots has shown impressive results in achieving complex tasks, and the huge potential to become a general robot platform~\citep{cheng2024tv, zhang2024wococo, cheng2024expressive, scironot2024humanoid}. It is a fundamental skill to support various humanoid motions, enabling them to navigate environments and perform tasks with agility and adaptability.
However, most current humanoid locomotion systems, although showing impressive results in motion-based control~\citep{he2024hover,2024exbody2,fu2024humanplus,cheng2024expressive} and mobile manipulation~\citep{lu2024pmp}, pay limited attention to producing versatile and controllable gait styles, leading to single, tedious, unextendable, and unconstrained movements.
Consider humans, we have versatile athletic abilities, such as running, jumping, and even hopping. Even when only walking, we can fine-tune our frequencies, strides, and foot heights. Bringing such versatility into humanoid locomotion is challenging, but it is the key to exploring the edge of humanoid robots' abilities.
\begin{figure*}[htbp]
\centering
\includegraphics[width=0.98\linewidth]{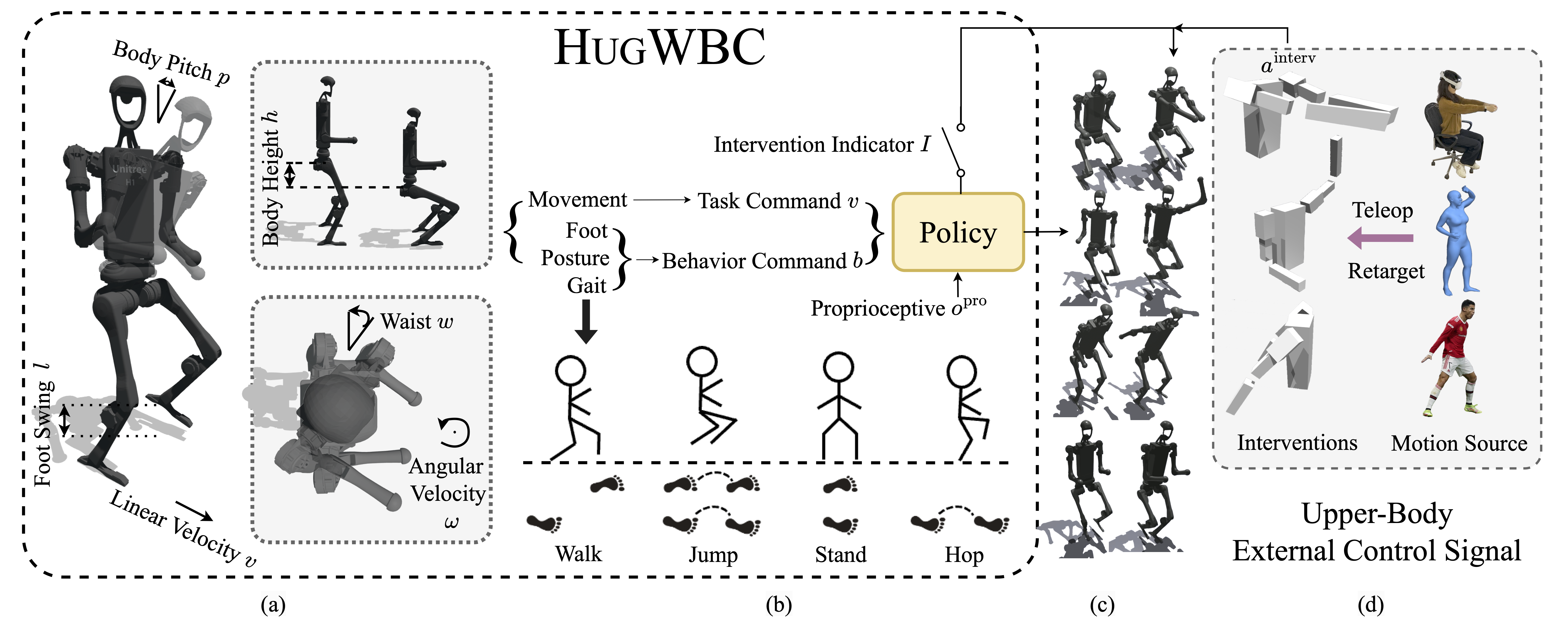}
\vspace{-10pt}
\caption{\small \textbf{Framework of \textsc{HugHBC}.} Illustration with the Unitree H1 robot. \textbf{a): Visualization of parts of commands}. The side view (left) highlights the linear velocity, foot swing height, and body pitch commands. The top-right view shows the angular velocity and waist yaw commands, and the bottom-right view shows the body height command. \textbf{b): Policy inputs/outputs.} The policy is provided with commands, proprioceptive observations, the intervention indicator, and outputs all joints of the robots. \textbf{c): Illustrations of four gaits on the robot without/with external intervention.} By default, the policy controls both the upper-body and the lower-body joints. \textbf{d): External control support.} Feasible external control signals can be seamlessly integrated into the robot's behavior without hurting locomotion performance.}
\label{fig:framework}
\vspace{-20pt}
\end{figure*}
To resolve the challenge and build a unified and general humanoid whole-body controller, in this work, we propose \our, namely, \textbf{H}umanoid's \textbf{U}nified and \textbf{G}eneral \textbf{W}hole-\textbf{B}ody \textbf{C}ontrol.
\our is designed for generating versatile locomotion with dynamic, customizable control, enabling the robot to perform gaits such as walking, standing, jumping, and hopping. Furthermore, \our provides the flexibility to adjust foot behavior parameters 
% stride length, removed
foot swing height and gait frequency, and allows combining posture parameters such as body height, waist rotation, and body pitch. 
To achieve this, \our includes a general command space designed for humanoid locomotion, along with advanced training techniques to learn versatile gaits within \textit{one single policy} (except the hopping gait) using reinforcement learning in simulation, which can be directly transferred onto real robots.

Positioned as a basic controller for humanoid robots to perform a wider range of tasks in diverse real-world scenarios, \our introduces intervention training and supports real-time external control signals of the upper body, like teleoperation, allowing for highly robust loco-manipulation, while maintaining precise locomotion control. An overview of the framework is illustrated in \fig{fig:framework}.

In experiments, we show \our preserves high tracking accuracy on eight different commands under four different gaits; we also ablate the improvement in stability and robustness of the upper body intervention training. We further provide a detailed analysis of how commands combination works, shedding light on the intricate relationships between these commands and how they can be leveraged to optimize movement performance.
Through this work, we aim to significantly broaden the scope of humanoid locomotion capabilities, pushing the boundaries of what is possible with current robotic systems.

The key contributions are summarized as follows:
\begin{itemize}[leftmargin=10pt]
    \item An extended general command space with advanced training techniques designed for versatile humanoid locomotion.
    \item Accurate tracking for eight different commands under four different gaits, using one policy for 3 of the 4 gaits.
    \item A basic humanoid controller that supports external upper-body intervention and enables a wider range of loco-manipulation tasks.
\end{itemize}
\section{Related Work}
\subsection{Model-Based Humanoid Controller}
% Trajectory optimization is a widely adopted instance of optimal control, which has been extensively applied in recent decades for the design of locomotion in humanoid systems and the generation of multi-contact gait patterns.

% Model Predictive Control (MPC) is a widely adopted approach within optimization-based techniques for humanoid robots. By utilizing either whole-body dynamics models~\citep{schultz2009modeling, koenemann2015whole} or simplified dynamics models, such as centroidal dynamics~\citep{orin2013centroidal, wensing2016improved} and the linear inverted pendulum (LIP) model~\citep{kajita2010biped}, researchers compute optimal trajectories by solving trajectory optimization (TO) problems~\citep{betts1998survey}. Some studies integrate contact mechanics, such as the zero-moment point (ZMP)~\citep{vukobratovic2004zero}, into simplified dynamic models~\citep{sugihara2009standing, wang2024online}, enhancing control strategies for stability and adaptability.

Controlling humanoid robots has become one of the most fascinating problems since decades ago, many researchers and engineers have built complicated systems and tried to solve them with model-based methods in a perspective of optimal control (OC)~\citep{2021ralFootstep, 2020rasRecedingHorizonPlanning, 2018troMulticontact, 2020troC-CROC, 2018Tower, 2020icraCrocoddyl, 2021troPatternGeneration}.
These works typically employ trajectory optimization with dynamic models of varying levels of complexity, such as the linear inverted pendulum model~\citep{kajita2010biped}, centroidal dynamics model~\citep{orin2013centroidal, wensing2016improved}, or full-body dynamics model~\citep{schultz2009modeling, 2015rasHRP-2humanoid, Xinjilefu}, to perform online optimization, or generate periodic motion control through the hybrid zero dynamics model~\citep{da2019combining,sreenath2011compliant, Hereid2016ICRA}. However, most of them can only generate motion based on predefined contact sequences. Even some have successfully incorporated online optimization to generate real-time motion sequences and contact schedules based on instant environmental feedback and user commands and run on humanoid robots in the real world~\citep{2019rasFootstep}, the nonlinear dynamics and multi-contact optimization of humanoid systems demand significant computational resources, making it challenging to meet real-time performance requirements.
A promising solution is to decouple the whole-body multi-contact optimization control problem into two subproblems: contact planning and motion optimization~\citep{2023troBiConMP, 2020icraCrocoddyl, AmesAaronDecopule}. The goal of the contact planning stage is to generate the desired multi-contact sequence for rich whole-body motion and gait control, including the order and position of both hand and foot contacts~\citep{2016rasmomentumdynamics, 2015rasHRP-2humanoid}. The motion optimization phase optimizes the robot motion trajectory based on the contact sequence. Although decoupling simplifies the problem, model-based approaches still rely on several assumptions, including perfect state estimation and flawless execution of planned movements. However, most assumptions no longer hold in the real world, and the dynamics model is not perfect to describe real robot systems, which results in poor robustness when applied in real environments.

\subsection{Learning-Based Humanoid Controller}
% Humanoid robot control remains a long-standing challenge due to the complexity arising from high degrees of freedom and low stability. 

Recent advancements in learning-based controllers have demonstrated the locomotion capability to go through rough terrains~\citep{scironot2024humanoid, rss2024denoisingworldmodel}, achieving smooth and efficient motions~\citep{chen2024learning}.
However, controllers relying on proprioceptive sensing must predict surrounding terrain through collision detection and swiftly adapt their motion, presenting significant challenges for inherently unstable humanoid robotic systems.
Some recent approaches incorporated depth maps or elevation maps into the policy observations, enabling impressive parkour tasks~\citep{zhuang2024humanoid,long2024learning}.
Some researchers have utilized chain-contact reward functions to learn jumping gaits for humanoid robots~\citep{zhang2024wococo}. 

Additionally, with the support of teleoperation systems for humanoid robots~\citep{cheng2024tv, fu2024mobile} and large-scale humanoid motion datasets~\citep{mahmood2019amass}, researchers have made progress in motion tracking and learned rich whole-body motion representations for humanoid robots.
Some studies focused on upper body tracking combined with maintaining balance in the lower body~\citep{cheng2024expressive}. Some others explored controlling whole-body joints in one policy, differing primarily in their control interfaces/command spaces: \citeauthor{he2024learning} tracked whole-body motion capture keypoints; \citeauthor{fu2024humanplus,2024exbody2} track retargeted joint position; \citeauthor{he2024omnih2o} tracked VR-based head and hands keypoints; \citeauthor{he2024hover} tracked all of these and propose a universal interface approach. Different from them, \citeauthor{lu2024pmp} decoupled the control interface, and combined an IK-based upper-body controller with a learning-based lower-body controller. The lower-body command space includes the task command and the pose command as used in this work, and they introduced the prior knowledge of upper-body movements to the lower-body policy to help its robustness. However, we show that without such a component, we can still construct a robust loco-manipulation controller.

We made several choices in this work: 1) we extend the command space beyond all of these previous works, by introducing additional behavior commands that control the foot and the gait; 2) we employ a learning-based controller to control whole-body joints (instead of only lower-body as in \citeauthor{lu2024pmp}) while supporting external controller (with IK or joint sequences) to take over upper-body joints, since upper-body and lower-body serves as different requirements. Accurate upper-body control is useful for tasks that require precision, while the robot should be robust to arbitrary upper-body intervention under any behavior.
% However, these approaches offer only limited interface implementations, constraining the flexibility of humanoid robots in performing advanced tasks. 
% Furthermore, current controllers exhibit monotonous motion patterns, lacking the ability to generate diverse humanoid gaits. 
% In contrast to previous works, our controller supports multiple motion modes and versatile whole-body control in response to user commands, while providing versatile interfaces for executing loco-manipulation tasks. OLD VERSION
% In contrast to previous works, \our enables multiple gait modes and offers versatile whole-body control in response to user commands. It also provides direct external control interfaces for upper body motion without any loss in tracking performance, eliminating the need for intermediate tracking policies in previous methods.

\section{Background}
\subsection{Humanoid Whole-Body Control}
To support various high-level functionalities and allow the humanoid robot to perform complicated tasks, a basic whole-body controller is essential.
Formally, given a set of continuous commands $\mathcal{C}$ and observations $\mathcal{O}$, our objective is to develop a control function that maps these inputs to appropriate control signals.
Model-based approaches represent one solution paradigm, typically decomposing the control function into planning and tracking modules~\citep{nolinearMPC2023, 2023troBiConMP}. The planning module generates optimal trajectories and contact sequences based on $\mathcal{O}$ and $\mathcal{C}$, while the tracking module translates these into control laws, specifying joint positions, velocities, and torques. However, these methods face computational challenges due to the complex dynamics of humanoid robots and the discrete nature of whole-body contact points. 
Learning-based methods offer an alternative approach by directly learning a policy function $a=\pi(o,c)$ that maps observations $\mathcal{O}$ and commands $\mathcal{C}$ to joint-level actions~\citep{2021BipedalPeriodicReward}. These actions typically represent offsets to target joint positions across three categories: upper-body, lower-body (legs), and hands. The final target position combines the nominal position with these learned offsets, which is then tracked using a proportional derivative (PD) controller with fixed gains.
% \subsection{}
% We formulate the locomotion problem as a Markov decision problem (MDP) in the discrete-time domain. At each time step t, the humanoid robot observes the state $x_t$ of the environment, and the policy $\pi_\theta$ generates a distribution of actions based on the state of the environment. The robot then executes the action $a_t$ sampled from the action distribution, and the environment transitions to the next state $x_{t+1}$ based on the state transition probability function $P \left( x_{t+1} | x_t, a_t \right)$ and returns a reward $r_{t}$. The goal of reinforcement learning is to find an optimal parameter to maximize the expected discounted return of future trajectories:
% \begin{equation}
%  J_{\theta}\left(\pi \right)  = \mathbb{E}_{\pi} \left[ {\sum\limits_{t = 0}^\infty  {{\gamma ^t}{r_t}} } \right],
% \end{equation}
% where $\gamma\in\left[0, 1\right)$ is discount factor. In order to enable a single policy to generate diverse behaviors, we designed a parameterized behavioral space $b_t$ and commands $c_t$ that simultaneously generates action distributions based on the state of the environment and the commands.

\subsection{Command Tracking as Reinforcement Learning}
To achieve a generalized and powerful whole-body control behavior for humanoid robots, we learn a policy function by constructing a command-tracking problem. In detail, we want the learned policy $\pi$ to control the robots to match the provided commands $c$. To this end, we use reinforcement learning (RL), where we define the reward functions $r$ typically by distances $d$ or similarities $s$  of the command $c$ and the observed robot state $s_c$ corresponding to that command:
\begin{equation}
\begin{aligned}
r(o, a, c) = -d(c, s_c) \text{ or } r(o, a, c) = s(c, s_c)~.
\end{aligned}
\end{equation}
Under the formulation of RL, the policy is trained to maximize the rewards, corresponding to matching these commands.

\subsection{Simulation Training and Real-World Transfer}
Many recent works, especially those of legged robots, take advantage of RL training a robust robot-control policy with a large set of parallel environments in simulation and directly deploying into the real world~\citep{cheng2024extreme,liu2024visual,he2024hover, sci2024dtc}.
Due to the partial observability of the real robot, whose onboard sensors can only provide limited and noisy observations, it is difficult to learn a deployable policy from them directly. 
Therefore, researchers have developed a set of sim-to-real techniques to resolve the challenge. Among them, one of the most commonly used techniques is asymmetric training~\citep{asyframework,nahrendra2023dreamwaq}, which is proposed as a one-stage solution for sim-to-real training.
% Among them, two of the most commonly used techniques are teacher-student training~\citep{lee2020learning} and asymmetric training~\citep{asyframework,ji2022concurrent}. Details of the algorithms are listed in \ap{ap:bk}.

% For the teacher-student training paradigm, a teacher policy with privileged states is first trained via RL (e.g., Proximal Policy Optimization (PPO)~\citep{schulman2017proximal}) in a highly efficient way, since the privileged states are low-dimensional and include clear task-related information. Then, a student policy is trained via online imitation learning (e.g., DAgger~\citep{ross2011reduction}) to mimic the actions of the teacher policy, which only observes the same partial and noisy observations as on real robots. 
% To mitigate the information loss during the distillation phase of the student policy, 
In this paper, we adopt an asymmetric actor-critic (AAC) framework proposed for quadruped locomotion~\citep{sci2023defrmable}.
In this framework, the critic network has access to all privileged information, while the actor network only receives data available from onboard sensors, with a separate encoder to estimate the key privileged information (\textit{e.g.}, the linear velocity, robot's body height, and robot's feet swing height). The training paradigm incorporates the RL objective (including a value loss $\mathcal{L}^{\text{value}}$ and a policy loss $\mathcal{L}^{\text{policy}}$) with an estimation loss \citep{nahrendra2023dreamwaq, liu2024skill, sci2023defrmable} $\mathcal{L}^{\text{est}}$ to train the encoder:
\begin{equation}
\begin{aligned}
    \mathcal L^{\text{AAC}} = \mathcal L^{\text{value}} + \lambda^{\text{policy}} \mathcal L^{\text{policy}} + \lambda^{\text{est}} \mathcal{L}^{\text{est}}
\end{aligned}
\end{equation}
In this work, we take AAC as our default training framework, but the proposed techniques are not limited to it.

\section{\our}
\subsection{A General Command-Space for Humanoid Locomotion}\label{sec:command-space}
% The versatile control of humanoid robots can be decoupled into the control of upper and lower body. 
% In this work, we are more focused on the diversity of lower body movements. 
% The control of the both arms is achieved through teleoperation.
% \begin{figure}[htbp]
%     \centering
%     \includegraphics[width=\linewidth]{imgs/CommandVisualize_V4.pdf}
%     \label{fig:CommandGait}
%     \vspace{-24pt}
%     \caption{\small \textbf{Visualization of parts of commands}. The side view (left) highlights the linear velocity, foot swing height, and body pitch commands. The top view (lower right) illustrates the angular velocity and waist roll commands. The body height command is shown in the comparison (upper right). Illustrated with Unitree H1 robot.}
% \end{figure}
We define the command space of the humanoid whole-body controller $\mathcal{C}=\mathcal{K}\times\mathcal{B}$ by two sets of commands, the task commands $\mathcal{K}$ and the behavior commands $\mathcal{B}$. The task commands determine a goal for the robot to reach, typically for movement, while the behavior commands construct the specific behavior pattern of the humanoid robots. In this work, we specify the task command as the target velocity $v_{t} \in \mathbb{R}^3$ (including the longitudinal and horizontal linear velocities $v_{t,x}$, $v_{t,y}$ and the angular velocity $\omega_t$) at each time step $t$. As for the behavior command, we define the behavior command $b_t$ as a vector:
\begin{equation}
\begin{aligned}
{\mathopen{\bBigg@{1.5}{[}}}          
         \underbrace{f_t,
         l_t}_{\text{foot}},
         \underbrace{h_t,
         p_t,
         w_t}_{\text{posture}},
         \underbrace{\psi_t,
         \phi_{t,1},
         \phi_{t,2},
         \phi_{t,\text{stance}}}_\text{gait}
     {\mathclose{\bBigg@{1.5}]}}~,
\end{aligned}
\end{equation}
where
$f_t \in \mathbb R$ is the gait frequency and $l_t \in \mathbb{R}$ is the maximum foot swing height, both of which can be explained as foot behaviors.
Besides, $h_t \in \mathbb{R}$ represents the body height, $p_t \in \mathbb{R}$ is the body pitch angle, and $w_t \in \mathbb{R}$ is the waist yaw rotation. These commands can be regarded as controlling the posture behavior.
% $h_t \in \left[1.4, 1.8\right]$ represents the body height, $p_t \in \left[-0.1, 0.8\right]$ is the body pitch angle, $w_t \in \left[-1.0, -1.0\right]$ is waist (yaw) rotation, and $s_t  \in \left[0.1, 0.4\right]$ is the maximum foot swing height.
% represents the humanoid robot’s squatting and bending movements

Beyond the commands above, we further introduce distinct gaits, such as walking, standing, jumping, and hopping. To do so, we refer to legged gait control~\cite{2021BipedalPeriodicReward, margolis2022walktheseways} and define $\phi_i \in [0, 1)$, $i = 1, 2$ as two 
% independent was removed because of phi_2 = phi_1 + psi in sometime
time-varying periodic phase variables to represent the humanoid gaits, on behalf of two legs (feet).
These two phase variables can either be set as constants, or be computed by the phase offset $\psi$ and the gait frequency $f_t \in \mathbb{R}$ at each time: 
\begin{equation}\label{eq:phase}
\begin{aligned}
& \phi_{t+1, 1} = \left ( \phi_{t,1} + f_t \times \text{d}t \right)~, \\
& \phi_{t+1, 2} = \left ( \phi_{t+1, 1} + \psi \right)~,
\end{aligned}
\end{equation}
where $\text{d}t$ is a discrete time step. When following the computation of Eq. \ref{eq:phase}, each $\phi_i$ loops in a range of $[0,1)$, resulting in repeated phase cycles. 
$\phi_{\text{stance}} \in [0, 1]$ is the duty cycle, which divides the gait cycle into two stages: stance (i.e., foot in contact with the ground) when $\phi_i < \phi_{\text{stance}}$, and swinging (i.e., foot in the air) otherwise. $f$ is the stepping frequency, determining the wall time of each gait cycle.

\begin{figure}
    \centering
    \includegraphics[width=\linewidth]{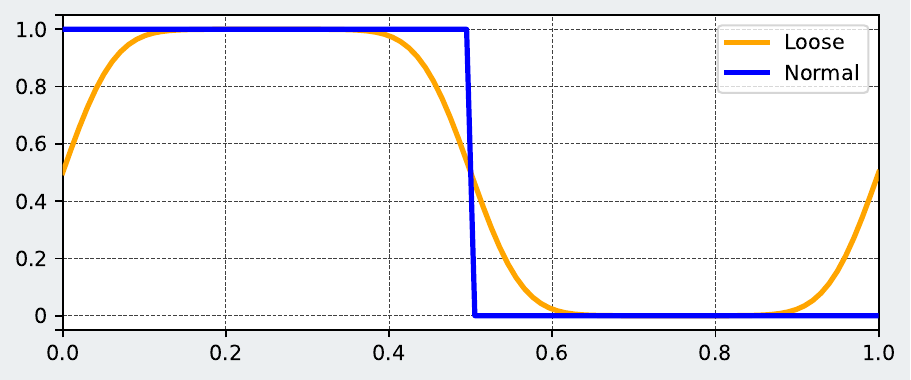}
    \caption{\small The expected contact probability function $C(\phi_{t,i})$ in the loose and normal formulation. The larger $C(\phi_{t,i})$, the higher the expectation of contact with the ground. The CDF of the normal distribution is introduced into the normal contact probability function to relax the constraint of the foot contact at the switching boundary, resulting in a smooth transition between the swing and the stance phase.}
    \label{fig:contact_curve}
    \vspace{-20pt}
\end{figure}

\noindent\textbf{Humanoid gait control.} 
\label{Humanoid gait control}
We consider four distinct standard gaits in this project, \textit{i.e.}, \textit{walking}, \textit{jumping}, \textit{standing}, \textit{hopping}\footnote{We note that \textit{running} can be further derived from the \textit{walking} gaits via high-velocity and small duty cycle commands, which promotes the prolonged flight of both feet.}
By constructing the behavior commands above, we can adjust the phase offset $\psi$, the duty cycle $\phi_{\text{stance}}$, and the phase variable $\phi_i$ for each leg to control the humanoid robots in versatile gaits.
In this work, we only consider standard gaits, so we set the phase offset $\psi = 0.5$ for \textit{walking} gaits \cite{2021BipedalPeriodicReward}, since the phase difference between the left and right foot is half a cycle;
Regarding \textit{jumping} gaits, the phase of the left and right foot is the same, thus, we set $\psi$ to 0. 
As for the \textit{standing} and the \textit{hopping} gaits, a certain foot of the robot is always in two states of contact or non-contact with the ground, which motivates a constant $\phi_i$ (resulting in constant contact probability of either 0 or 1, and $\psi$ is not working). In particular, for the standing gait, we set $\phi_i=0.25$ for both feet; and for the hopping gait, $\phi_i=0.75$ for the flying leg, and $\phi_i$ of the other leg steps with frequency $f$.
% Due to the rolling change of phase variables $\phi_i$ leading to a continuous transformation of the contact state between the foot and the ground, these modes are rather difficult to generate. 
% To resolve this challenge, we randomly sample the $\phi_i$ as a constant value for learning the standing and hopping gaits. 
% For the standing gait, the $\phi_i$ of both feet is set to 0. The $\phi_i$ of the flying leg is set to 0.75 and the $\phi_i$ of another leg steps with frequency $f$ for hopping gait. 
The $\phi_{\text{stance}}$ determines the time ratio of stance and swinging during a gait cycle, and a smaller $\phi_{\text{stance}}$ will promote longer leg flight time. 
To represent a smooth switch between stance and swinging, 
we introduce the expected contact probability function $C(\phi_{t,i})$ for leg $i\in\{1,2\}$ at each time step $t$ as:
\begin{align}
% \begin{aligned}
\begin{split}
C(\phi_{t,i}) &= \Phi(\bar{\phi}_{t,i} / \sigma) [1 - \Phi((\bar{\phi}_{t,i} - 0.5)/\sigma)] \\
& \quad + \Phi((\bar{\phi}_{t,i}- 1)/\sigma) [1 - \Phi((\bar{\phi}_{t,i} - 1.5)/\sigma)]~,
\end{split}
\\
\bar{\phi_i} &= \left\{\begin{aligned}
&0.5 \times \frac{\phi_i}{\phi_\text{stance}}, 
&\phi_i < \phi_\text{stance} \\ 
& 0.5 + 0.5 \times \frac{\phi_i - \phi_\text{stance}}{1 - \phi_\text{stance}},   &\phi_i \ge \phi_{\text{stance}}
\end{aligned}\right.~,\label{eq:bar_phi}
% \end{aligned}
\end{align}
%OLD VERSION
% \begin{align}
% % \begin{aligned}
% \begin{split}
% C^t_i(\phi_i) &= \Phi_i(\bar{\phi_i}) [1 - \Phi_i(\bar{\phi_i} - 0.5)] \\
% &+ \Phi_i(\bar{\phi_i}- 1) [1 - \Phi_i(\bar{\phi_i} - 1.5)]~,
% \end{split}
% \\
% \bar{\phi_i} &= \left\{\begin{aligned}
% &0.5 \times \frac{\phi_i}{\phi_{stance}}, 
% &\phi_i < \phi_{stance} \\ 
% & 0.5 + 0.5 \times \frac{\phi_i - \phi_{stance}}{1 - \phi_{stance}},   &\phi_i > \phi_{stance}
% \end{aligned}\right.~,\label{eq:bar_phi}
% % \end{aligned}
% \end{align}
where $\bar{\phi_i} \in [ 0, 1 ]$ is a homogenized phase variable that maps the $\phi_i$ of the stance and swinging phases to intervals $[0, 0.5]$ and $[0.5, 1]$ according to $\phi_{\text{stance}}$, as computed in \eq{eq:bar_phi}. $\Phi(\cdot)$ is the cumulative distribution function (CDF) of the standard normal distribution $N(0, 1)$. The standard deviation $\sigma$ allows for the relaxation of switching points ($\bar{\phi_i} = 0, 0.5$) to switching interval ($\bar{\phi_i} \in [-3 \sigma, 3 \sigma], [0.5 -3 \sigma, 0.5 + 3 \sigma]$)~(see \fig{fig:contact_curve} for a detailed explanation). 
Intuitively, $C(\phi_{t,i})$ is the probability of leg $i$ coming into contact with the ground.
As one may notice, when $\bar{\phi}_{t,i} \in [0, 0.5]$, the first term of $C(\phi_{t,i})$ is dominant; otherwise, the second term becomes dominant.
In this work, we set a constant $\phi_{\text{stance}}=0.5$ for all supported gaits in all time steps, which means half-time stance/swinging during one cycle.

We highlight that \our trained \textit{one single policy} for the standing, walking, and jumping gaits, and an independent policy for the hopping gait.

\subsection{Detailed Observation} 
\label{sec:OBS}
In our asymmetric actor-critic framework, the observation for the critic network $o^{V}_t$ obtains all information related to the environment,
including proprioceptive observations $o_t^{\text{pro}}$, privileged observations $o_t^{\text{pri}}$, terrain observations $o_t^{\text{ter}}$, commands $c_t$, and an indicator signal $I(t)$. Regarding the actor network, its available observation $o_t^{\pi}$ only contains history of proprioceptive observations within last $k$ steps $o_t^{\text{his}} = (o_{t-k+1}^{\text{pro}}, \dots, o_t^{\text{pro}})$, commands $c_t$, and the indicator signal $I(t)$.
The proprioceptive $o_t^{\text{pro}} \in \mathbb{R}^{63}$ consists of angular velocity and gravity projection in the robot's base frame, joint position, joint velocity, and previous policy output $a_{t-1}$. 
The privileged observations $o_t^{\text{pri}} \in \mathbb{R}^{24}$ contain the linear velocity, the base height error, foot clearance, friction coefficient of the ground, feet contact forces, and collision detection of the link (trunk, hip, thigh, shank, shoulder, and arm). The terrain observations $o_t^{\text{ter}} \in \mathbb{R}^{221}$ are samplings of terrain elevation points around the robot.

\begin{figure}[t]
    \centering
    \includegraphics[width=0.98\linewidth]{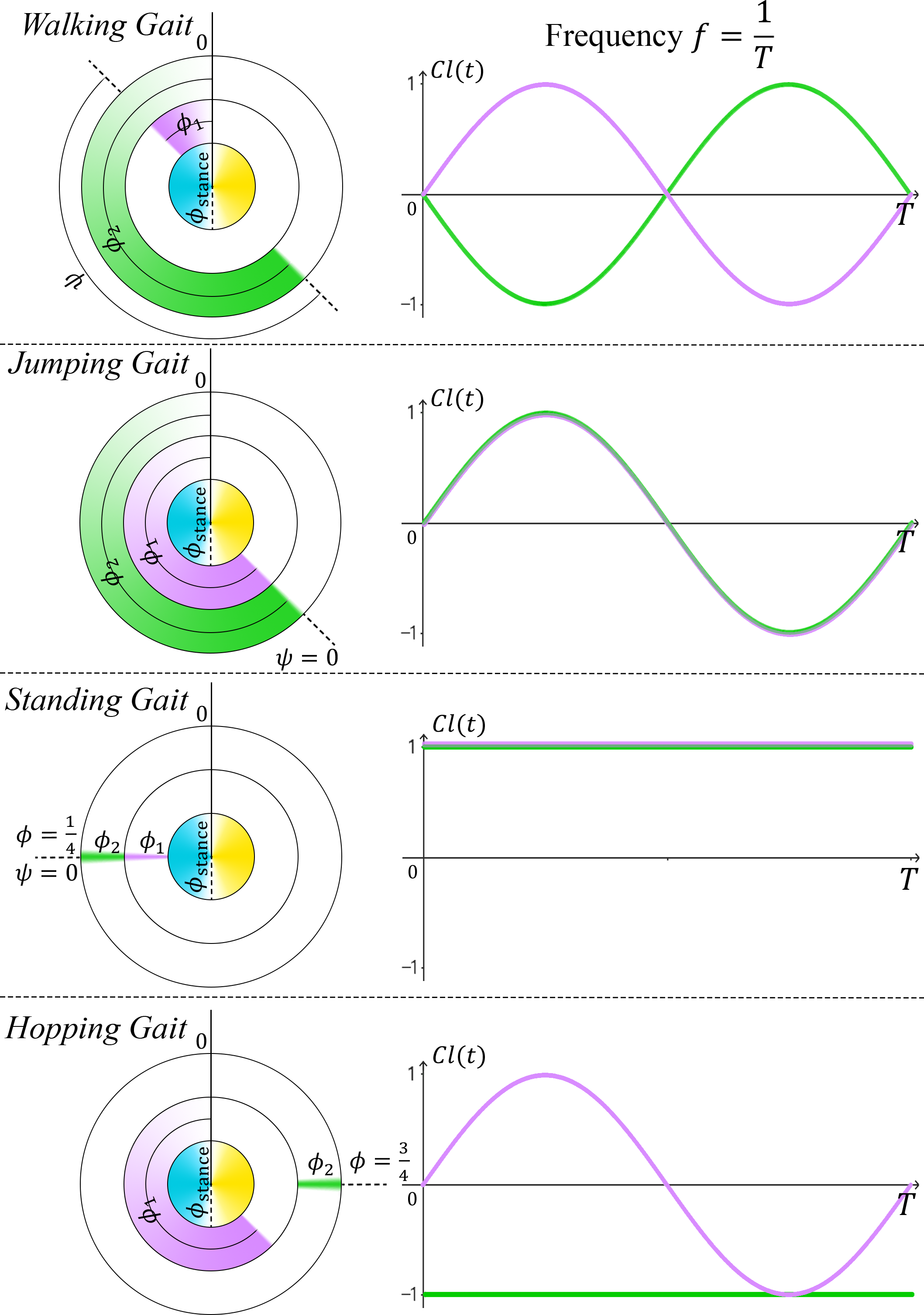}
    \vspace{-6pt}
    \caption{\small \textbf{Phase variables and clock functions under different gaits.} \textbf{Left}: The \textcolor{myPurple}{purple} ring represents the phase variable $\phi_1$ for the left foot, while the \textcolor{myGreen}{green} ring represents the phase variable $\phi_2$ for the right foot. $\psi$ is the phase offset from $\phi_1$ to $\phi_2$. The dividing phase between stance (marked in \textcolor{myBlue}{blue}) and swing (marked in \textcolor{myYellow}{yellow}) is the duty cycle $\phi_{\text{stance}} (0.5)$. \textbf{Right}: The \textcolor{myPurple}{purple} line depicts the clock function $Cl_L(t)$ for the left foot over a cycle, while the \textcolor{myGreen}{green} line represents the clock function $Cl_R (t)$ for the right foot over a cycle. 
    }
    \label{fig:GaitPhiPsiClock}
    \vspace{-10pt}
\end{figure}

\noindent\textbf{Commands.}
The commands $c_t=[v_t, \tilde{b}_t]$ includes the task command (\textit{i.e.}, target velocity $v_t$ in this work) and the extended behavior command $\tilde{b}_t \in \mathbb{R}^{9}$, 
% where we extend the behavior command $b_t$ defined above with two additional terms $[\sin\left(2\pi\phi_1^t\right), \sin\left(2\pi\phi_2^t\right)]$ representing the contact of both feet. 
where we extend the behavior command $b_t$ defined above through replacing the phase variables $\phi_i,~ i= 1, 2$ with two additional clock functions $[Cl_L(t), Cl_R(t)]=[\sin\left(2\pi\bar\phi_{t,1}\right), \sin\left(2\pi\bar\phi_{t,2}\right)]$ representing the contact of both feet, where $\bar \phi_{t,i},~ i=1,2$ are the homogenized phase variables defined in \eq{eq:bar_phi}.
Note that the sine function $\sin\left(2\pi\bar\phi_{t,i}\right), i= 1, 2$ is a gait cycle contact indicator function, designed for a smoother transition between swinging and stance phases. An illustrative explanation of the phase variables and the clock function is shown in \fig{fig:GaitPhiPsiClock}.  

\noindent\textbf{External upper-body control.}
We want to build a general humanoid whole-body controller that also supports external upper-body control (e.g., teleoperation). Thereafter, we introduce a binary indicator function $I(t)$ to identify whether an external upper-body controller is involved.
If there is no external upper-body control signal involved, the upper-body joints are controlled by our developed whole-body controller by default, which swings the arms naturally.

\subsection{Reward Design for Policy Learning}
Our humanoid whole-body controller is obtained through an asymmetric actor-critic training paradigm via reinforcement learning (RL). To learn a policy with general and diverse behaviors, we design a set of reward functions, which mainly consist of three parts: task rewards, behavior rewards, and regularization rewards. The details of the rewards are concluded in \tb{tab:rewards}.

\begin{table}[t]
\setlength{\abovecaptionskip}{0.cm}
\setlength{\belowcaptionskip}{-0.cm}
\caption{\small \textbf{Reward definitions used in \our.}}
\resizebox{\linewidth}{!}{
\centering
\label{tab:rewards}
\begin{tabular}{l|l|l}
\toprule
\multicolumn{1}{c|}{Term} & \multicolumn{1}{c|}{Definition}                                                                                                                                                               & \multicolumn{1}{c}{Weight} \\ \midrule
\multicolumn{3}{c}{Task Reward}                                                                                                                                                                                                                    \\ \midrule
Linear Velocity Tracking    & $\exp\left ( -\left \| v_{\text{xy}}^\text{target}-v_{\text{xy}} \right \|^2 / 0.2 \right) $                                                                                                                         & 2                        \\
Angular Velocity Tracking    & $\exp\left ( -\left \| \omega_{\text{z}}^\text{target}-\omega_{\text{z}} \right \|^2 / 0.2 \right) $                                                                                                                 & 2                        \\ \midrule
\multicolumn{3}{c}{Behavior Reward}                                                                                                                                                                                                                \\ \midrule
Body Height Tracking     & $\left \| h^\text{target}-h \right \|^2$                                                                                                                                                     & -40                        \\
Body Pitch Tracking      & $\left \| p^\text{target}-p \right \|^2$                                                                                                                                                     & -10                        \\
Waist Yaw Tracking       & $\left \| w^\text{target}-w \right \|^2$                                                                                                                                                     & -2                         \\
Foot Swing Tracking & $\sum_i [1-C(\phi_i)]\left \| l^{\text{target},i} - l^{\text{foot},i} \right \|^2$                                                                                                                       & -30                        \\
Contact-Swing Tracking   & \makecell[l]{$
-\sum_i [1-C(\phi_i)]\left[1-\exp(\left \|  f^{\text{foot}, i} \right \|^2 / 50)\right]$ \\
$~~~~~~~~~~~ -C(\phi_i)\left[1-\exp\left(\| v^{\text{foot}, i}_{xy} \|^2 / 5\right)\right]$} & -2                         \\ \midrule
\multicolumn{3}{c}{Regularization Reward}                                                                                                                                                                                                                 \\ \midrule
R-P Angular Velocity      & $\left \| \omega_{\text{xy}} \right \|^2$                                                                                                                                                       & -0.5                       \\
Vertical Body Movement            & $\left \| v_{\text{z}} \right \|^2$                                                                                                                                                           & -0.1                       \\
Feet Slip                & $1- \sum_i \exp\left(-\left \| v_{\text{xy}}^\text{foot,i} \right \|^2\right)$                                                                                                                                           & -0.2                       \\
Action Rate              & $\left \| a_t - a_{t-1} \right \|^2$                                                                                                                                                       & -0.01                      \\
Action Smoothness        & $\left \| a_{t-2}-2a_{t-1}+a_t \right \|^2$                                                                                                                                                & -0.01                      \\
Joint Torque             & $\left \| \tau \right \|^2$                                                                                                                                                            & -5e-6                      \\
Joint Acceleration       & $\left \| \ddot{q} \right \|^2$                                                                                                                                                            & -2.5e-7                    \\
Upper Joint Deviation    & $\left \| q_\text{upper} - q^\text{nominal}_{\text{upper}} \right \|^2$                                                                                                                                             & -0.5                       \\
Hip Joint Deviation      & $\left \| q_{\text{hip},\text{xz}} - q^\text{nominal}_{\text{hip}, \text{xz}} \right \|^2$             
& -2                         \\
Feet Symmetry         & $\mathds{1} [\bar \phi_1 = \bar \phi_2] \left \| p_\text{foot,0}^{\text{xz}}-p_\text{foot,1}^{\text{xz}} \right \|^2$                                                                                                        & -5                        \\
Termination & $\mathds{1} [\text{Early Terminate}]$ & -200 \\ \bottomrule
\end{tabular}}
\vspace{-6pt}
\end{table}

The \textit{task} rewards are meant to track any task command $k$. In this work, it is the target velocity $v$, including the linear and angular velocities. 
% The \textit{regularization} rewards are items for regularizing the behaviors during training.
The \textit{regularization} rewards take into account the performance of physical hardware and impose constraints on the smoothness and safety of the locomotion.
These are commonly used in previous works~\cite{rudin2022learning}.
In this work, since we want to build a general whole-body controller to support versatile locomotion behaviors for humanoid robots, we introduce a set of \textit{behavior} rewards to encourage the robots to track any behavioral commands $b$, shown below.

For most behavior commands, including body height $h$, body pitch $p$, and waist rotation $w$, we simply formulated the rewards with mean squared error (MSE):
\begin{equation}
\begin{aligned}
r_t^{\text{cmd}} = \left \| e_t^{\text{target}} - e_t^{\text{cmd}} \right \|^2~.
\end{aligned}
\end{equation}
Beyond these simple tracking rewards, we further introduce periodic contact-swing rewards $r_t^{\text{contact}}$~\citep{2021BipedalPeriodicReward, margolis2022walktheseways} and the foot trajectory rewards $r_t^{\text{traj}}$ to help generate complicated gaits.

The periodic contact-swing reward $r_t^{\text{contact}}$ is designed for precise adjustments between swinging and stance in different gaits, according to $\phi_i$. 
Since humanoid gaits can be expressed as different combinations of contact sequences, like foot contact forces and velocities, we define the periodic contact-swing rewards $r^{\text{contact}}_t$ over them to generate desired contact patterns.
Based on $C(\phi_{t,i})$ defined as \eq{eq:bar_phi}, we then construct the periodic contact-swing rewards $r_t^{\text{contact}}$ to encourage humanoid robots to learn specific contact modes and generate various humanoid gaits:
\begin{equation}
\begin{aligned}
r_t^{\text{contact}} = & -\sum_{i=1}^{2} [1-C(\phi_{t,i})]\left[1-\exp\left(\left \|  f^{\text{foot}, i}_t \right \|^2 / \sigma_{cf} \right)\right] \\
& -\sum_{i=1}^{2} C(\phi_{t,i})\left[1-\exp\left(\| v^{\text{foot}, i}_{t, xy} \|^2 / \sigma_{cv} \right)\right]~,
\end{aligned}
\end{equation}
where $f_t^{\text{foot}, i}$ denotes the foot contact force and $v^{\text{foot}, i}_{t, xy}$ is the foot velocity. $\sigma_{cf}$ and $\sigma_{cv}$ are hyperparameters, fine-tuned according to the range of previous work \citep{margolis2022walktheseways} (We set the value as $\sigma_{cf}=50$, $\sigma_{cf}=5$). Note that during the stance phase, this reward function penalizes the foot velocities and ignores the foot contact force; on the other hand, during the swing phase, it penalizes the foot contact force and ignores the foot velocity.

Except for the contact in gait control, we also require the foot to smoothly reach the highest point and fall down, ensuring a precise and controllable swing. We introduce the foot trajectory reward $r_t^{\text{swing}}$ to achieve this:
\begin{equation}\label{eq:foot_traj}
\begin{aligned}
r_t^{\text{swing}} = & \sum_{i=1}^2 [1-C(\phi_{t,i})]\left \| l_t^{\text{target}, i} - l_t^{\text{foot},i} \right \|^2~.
\end{aligned}
\end{equation}
Note that in \eq{eq:foot_traj}, $l_t^{\text{foot},i}$ denotes the actual swing height of foot $i$, $C(\phi_{t,i})$ is the expected contact probability function. 
$l_t^{\text{target}, i}$ is the target swing height, derived from a desired foot trajectory, as discussed below.

A desired foot trajectory should typically require the fulfillment of three key criteria: 
1) zero foot velocity and acceleration during the stance phase; 2) zero foot velocity and acceleration at the end of the swing phase; and 3) continuity of both foot velocity and acceleration during the transition between the two phases. This is beneficial for enhancing motion stability and reducing energy consumption. 
In this work, we follow the experience in robot kinetics and quadruped robots~\citep{2016RasHermiteSplines, LocManMPC2021RAL}, and incorporate the quintic polynomial trajectory to compute the target swing height $l_t^{\text{target}, i}$ at each control step: % The Bezier and polynomial curves are commonly employed in foot trajectory planners for quadruped robots.
\begin{equation}
\begin{aligned}
l_t^{\text{target},i} = \left\{\begin{aligned} 
&l_t \sum_{k=0}^{5} a_k \left(0.25 - \left| \bar{\phi}_{t,i} -0.75 \right| \right)^k   , &\bar{\phi}_{t,i} >0.5\\ 
& 0, & \bar{\phi}_{t, i} < 0.5 \end{aligned}\right.~.
\end{aligned}
\label{eq:ltarget}
\end{equation}
Here $l_t$ is the foot swing height command, and the polynomial coefficient $a_k$ is determined based on the homogenized phase variable $\bar{\phi_i}$, as well as the boundary conditions of swing position, velocity, and acceleration. A detailed explanation of the calculation process is provided in the \ap{ap:Foot Target}. Note that \eq{eq:ltarget} only defines the target trajectory in the $z$-axis. On natural terrains, the robot's precise foothold planning is not required. As for swing trajectories in the $x$-axis and the $y$-axis, which determines the stride, they can be computed based on the gait frequency $f$ and the velocities $v,\omega$~\citep{gehring2013control,linearmpc2018cheetah}. 
% \yufei{The complete foot trajectory of a robot includes trajectories in the x, y, and z directions. Equation (10) only represents the trajectory in the z direction, and x, y determines the robot's stride. However, in this paper, we introduce gait frequency and velocity, which jointly determine the length of the trajectory in the x and y directions. Therefore, there is no need to set a reward. Do we need the derivation formulas for x, y, and stride length, as well as the reason for not displaying the reward for constructing the x and y trajectories?}

\subsection{Mirror Function and Symmetry Loss}
Natural and symmetrical motion behavior is gradually mastered by humans through acquired learning, due to its inherent elegance and efficiency in minimizing energy expenditure. Humanoid robots, with highly biomimetic mechanisms, also have symmetrical structural features.
However, without prior knowledge, the symmetrical morphology information is difficult to be explored by the policy, especially for policies that generate diverse behaviors. This makes the initial exploration much more difficult, making the policy easily fall into local optima and leading to unnatural movements.
To leverage the advantage of this morphological symmetry and inspired by \cite{symmetry2018tog}, we proposed the mirror function $\mathcal{F} \left( \cdot \right)$ for a humanoid robot to encourage the policy to generate symmetric and natural motion. Under such a symmetrical structure, ideally, the policy output should satisfy:
\begin{equation}
\begin{aligned}
\pi (o_t^{\pi}) = \mathcal{F}_a (\pi(\mathcal{F}_o (o_t^{\pi}))) ~.
\end{aligned}
\end{equation}
Intuitively, the mirror function produces a mirror output symmetric to the X-Z plane.
Here $\mathcal{F}_a$ and $\mathcal{F}_o$ are called \textit{action mirror function} and \textit{observation mirror function}, respectively, which map actions and observations to their mirrored version. Derived from these symmetric functions, we define a symmetry loss function $\mathcal{L}_\text{sym}$. The policy learning objective for controlling robots with symmetrical structures can be written as:
\begin{equation}\label{eq:sym-obj}
% \begin{aligned} %OLD VERSION
% \mathcal{L}(\pi)_\text{sym} = \mathbb{E}_{\tau \thicksim {\pi}} [\sum_{t=0}^{\infty}\Vert \pi \left({o_t^{\pi}} \right) - \mathcal{F}_{a} \left( \pi \left( \mathcal{F}_{o}  \left( {o_t^{\pi}}\right) \right) \right) \Vert^2]~,
% \end{aligned}
\begin{aligned}
\mathcal{L}_\text{sym} = \sum_t\left\| \pi \left({o^{\pi}_t} \right) - \mathcal{F}_{a} \left( \pi \left( \mathcal{F}_{o}  \left( {o^{\pi}_t}\right) \right) \right) \right\|^2~,
\end{aligned}
\end{equation}
The $\mathcal{L}_\text{sym}$ is independent of the RL objective, making it easy to extend to different RL algorithms. 
It is worth noting that the symmetric loss function is in fact encouraging symmetric actions on symmetric states (and commands), and it can be utilized for behaviors from symmetric ones (like walking and jumping) to asymmetric ones, such as hopping gaits, where hopping with the left foot is symmetric to hopping with the right one.

\noindent\textbf{Overall training objective.}
\our adopt an asymmetric actor-critic framework~\citep{nahrendra2023dreamwaq}, taking PPO~\citep{schulman2017proximal} as the RL algorithm to train the whole-body policy. Therefore, the total training objective can be written as:
\begin{equation}\label{eq:total-obj}
\mathcal{L} = \mathcal{L}_\text{AAC} + \beta \mathcal{L}_\text{sym}, \end{equation}
where $\beta$ is a weight coefficient to balance between minimizing the RL objective and symmetry gait (we simply set $\beta$ = 0.5 in our practice). 
We implemented a critic network, an actor network, along with the privileged encoder, all as Multi-Layer Perceptrons (MLPs). The actor network, combined with the encoder, can be directly deployed onto the real robot at a control frequency of 50 Hz. The sampled trajectories have a maximum length of 1000 timesteps, and the termination conditions include trunk collision with the ground or other links, as well as large body inclinations.

\subsection{External Upper-Body Intervention Training}
\label{sec:intervention}
So far we learned a whole-body controller, which controls the upper and lower body jointly.
Nevertheless, the goal of this work is not a controller specifically designed for locomotion tasks, but to build a unified and general humanoid controller that can serve as a basic support for loco-manipulation tasks. In other words, our controller should also support flexible and precise control of the upper body (arm and hands). Different from some previous works~\citep{he2024omnih2o,he2024hover} that augment the command space with upper body commands (\textit{e.g.}, arm joint positions), we consider decoupling the upper body control as external control intervention by teleoperation signals~\citep{cheng2024tv,lu2024pmp} or retargeted motion joints~\citep{cheng2024expressive,2024exbody2}, while not affecting the lower-body gaits, due to their high control precision.
Our solution is sampling alternative actions to replace the upper-body actions produced by the whole-body policy during training, making the policy robust to any intervention.

\noindent\textbf{Switching between whole-body control and intervention.}
Denote $I(t)$ a binary indicator function for whether the external control signal intervenes at each time step $t$, we assign a small probability of $p$ ($p=0.005$ in this work) to reverse $I(t)$.
This leads the expected length of a continuous sequence without changing the upper-body control mode to be $\sum_{n=1}^{\infty}np(1-p)^n=\frac{1-p}{p}$,
ensuring infrequently switching between two control modes and most of the trajectories are either long sequence of whole-body controlling or intervention, preventing rapid switches.
% By this way, the expectation of a continuous sequence length with or without changing upper-body states is $(1-p)/p$, taking into consideration of three situations: long sequence of whole-body controlling, long sequence of intervenes and the switching between them.

% two sources: 1) the filtered motion capture dataset AMASS~\citep{he2024omnih2o}, which is retargeted from the SMPL model to the morphology of our robot;
% ~\citep{smpl, expressive_humanoid}; NOT FOR THE NEWEST VERSION
% During training, we mix the two sources in a ratio of 0.7 for the noises to 0.3 for the motion dataset.
\noindent\textbf{Intervention sampling.}
The intervened actions of the humanoid upper body are sampled from uniform noises, which introduce the potential for collisions with the body, simulating erroneous operations during external intervention. 
\begin{figure}[t]
\centering
\includegraphics[width=\linewidth]{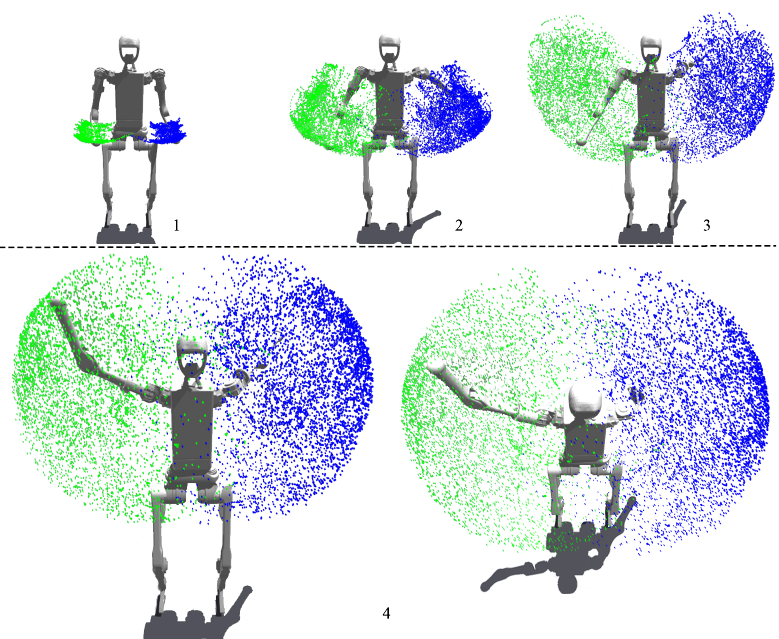}
\vspace{-8pt}
\caption{\small \textbf{Intervention noise curriculum.} We illustrate sampled noise by visualizing the hand positions relative to the visualized robot hand joints. \textbf{Top 1-3}: Noise samples of three curriculum stages with noise levels ranging from small to large. These noises are only relative to the robot hand joints as visualized in the figures. \textbf{Bottom 4}: Front and top views of the noise samples from the final noise curriculum.}
\label{Fig: Noise}
\vspace{-10pt}
\end{figure}

\noindent\textbf{Noise intervention interpolation.}
To prevent meaningless jitters caused by noise intervention sampling, the intervention action $a_{\text{noise}}^{\text{targ}}$ is randomly sampled in the action space every $t_\text{interval} = 90$ time steps. 
During the first two third time steps in the interval, linear interpolation is applied to smoothly transition the intervention joint positions from the initial pose $a_{\text{noise}}^{\text{init}}$ to the target pose $a_{\text{noise}}^{\text{targ}}$, while the target intervention action is maintained for the remaining time steps.
\begin{equation}
    \begin{aligned}
        &a^{\text{interv}}_{\text{noise},t} = (1-r)a_{\text{noise}}^{\text{init}} + r a_{\text{noise}}^{\text{targ}} ~, \\
        &r = \min \left(1, \frac 3 2\frac{t-t_0}{t_{\text{interval}}} \right) ~.
    \end{aligned}
\end{equation}
In this equation, $r$ is the ratio for the linear interpolation and $t_0$ is the sampled time.
% Old Version
% And the intervention action going to be executed is 
% \begin{equation}
%     a^{\text{interv}}_{\text{noise}, t} = \text{clip}(a_t^{\text{policy}}+\epsilon, q_t - r, q_t + r)~,
% \end{equation}
% where $a_t^{\text{policy}}$ is the output of the whole-body control policy, $\epsilon \sim U(-1,1)$ is the noise, $q_t$ is the joint position of upper body at time step $t$, $r$ is the maximum action rate. 

\noindent\textbf{Reward mask.}
When the intervention is involved, we mask the regularization rewards of the upper body during training, in order to eliminate the potential conflict of the policy output that tries to take over the upper body.

\noindent\textbf{Noise curriculum.}
The replaced intervention action $a_t^{\text{interv}}$ is gradually transited from the policy action $a_t$ to the sampled noise $a^{\text{interv}}_{\text{noise}, t}$:
\begin{equation}
    a_t^{\text{interv}} = \alpha a^{\text{interv}}_{\text{noise}, t} + (1-\alpha) a_t~,
    \label{eq:noisecurr}
\end{equation}
where the smoothing factor $\alpha$ increases per the progression of the intervention curriculum. In detail, $\alpha$ increases by 0.01 when both the linear and angular velocity tracking rewards exceed predefined thresholds. Conversely, if either of the velocity rewards fails to reach two-thirds of these thresholds, $\alpha$ is decreased by 0.01. The noise curriculum is illustrated in \fig{Fig: Noise}.

\subsection{Curriculum Learning}
Directly learning a diverse policy from manual rewards presents significant challenges due to the simultaneous optimization and exploration of multiple objectives. We thereby propose a curriculum learning approach to improve training efficiency. In particular, we split two distinct parallel robot training groups: an ``agile group'', tasked with learning high-speed, agile locomotion, and an ``intervention group'', focused on developing a control policy for managing external upper-body interventions.
At the beginning of training, each group of robots randomly samples one specific gait from four humanoid gaits, \textit{i.e.}, standing, walking, jumping, hopping. The remaining behavioral commands $(f_t, l_t, h_t, p_t, w_t)$ and the task commands $v_t$ are uniformly sampled from the specified ranges, which can be further referred to in \tb{tab:commands}. 
% \wentao{If there exists a better and simpler way to say the initial range than the table? }
Following \cite{wjzamp}, we employ a terrain curriculum for both groups, which consists of continuous rough terrain. Once the robot successfully masters the most challenging terrain, we keep that terrain and initiate an intervention noise curriculum and a speed curriculum simultaneously.
On the one hand, the speed curriculum only works for the agile group, meant to learn high agility, which gradually increases the speed commands $v_t$ following a grid adaptive curriculum strategy~\citep{margolis2022rapid}.
On the other hand, the intervention noise curriculum as described in \se{sec:intervention} works for the intervention group, focused on working with arbitrary upper-body intervention signals.

\section{Simulations and Experiment}
In this section, we conduct comprehensive experiments in both simulation and the real-world robot to address the following questions:
\begin{itemize}[leftmargin=*]
    \item \textbf{Q1(Sim)}: How does the \our policy perform in tracking across different commands?
    \item  \textbf{Q2(Sim)}: How to reasonably combine various commands in the general command space? % Command Analysis
    \item \textbf{Q3(Sim)}: How does large-scale noise intervention training help in policy robustness? % Ablation Study
    \item \textbf{Q4(Real)}: How does \our behave in the real world? % Real World Demo
\end{itemize}

\noindent\textbf{Robot and Simulator.} 
Our main experiments in this paper are conducted on the Unitree H1 robot, which has 19 Degrees of Freedom (DOF) in total, including 
two 3-DOF shoulder joints, two elbow joints, one waist joint, two 3-DOF hip joints, two knee joints, and two ankle joints.
The simulation training is based on the NVIDIA IsaacGym simulator~\citep{makoviychuk2021isaac}. It takes 16 hours on a single RTX 4090 GPU to train one policy.

\noindent\textbf{Command analysis principle and metric.}
One of the main contributions of this paper is an extended and general command space for humanoid robots. Therefore, we pay much attention to command analysis (regarding Q1 and Q2). This includes analysis of single command tracking errors, along with the combination of different commands under different gaits.
% we categorize the commands into three groups: \emph{movement}, \emph{foot}, and \emph{posture}. The \emph{movement} commands include the linear velocity and angular velocity, forming the foundational locomotion commands and are considered the most critical aspect of the tasks. The \emph{foot} commands include the gait frequency and foot swing height, representing the mode of leg movement. The \emph{posture} commands include body height, body pitch and waist yaw, which determine the desired body posture.
For analysis, we evaluate the averaged episodic command tracking error (denoted as $E_\text{cmd}$), which measures the discrepancy between the actual robot states and the command space using $L_1$ norm.
% The tracking error is measured in units of $m/s$, $rad/s$, $Hz$, $m$, and $rad$, corresponding to linear velocity, angular velocity, frequency, position, and rotation, respectively.
All commands are uniformly sampled within a pre-defined command range, as shown in \tb{tab:commands}\footnote{Note that the hopping gait keeps a different command range, due to its asymmetric type of motion. More details can be referred to \ap{ap:Hopping}.}.

\begin{table}[t]
\setlength{\abovecaptionskip}{0.cm}
\setlength{\belowcaptionskip}{-0.cm}
    \centering
    \caption{\small \textbf{Command ranges.} Ranges of curriculum starting, finishing, and default values of commands, for all gaits except hopping.} 
    \label{tab:commands}
    \begin{tabular}{c|cc|cc}
\toprule
Group                     & Term                      & Default & \begin{tabular}[c]{@{}c@{}}Initial\\ Range\end{tabular} & \multicolumn{1}{c}{\begin{tabular}[c]{@{}c@{}}Finishing\\ Range\end{tabular}} \\ \midrule
\multirow{3}{*}{\makecell{Task\\Commands}} & $v_x$     & 0       & $[-0.6, 0.6]$                                           & $[-0.6, 2.0]$                                                                 \\
                          & $v_y$     & 0       & $[-0.6, 0.6]$ & $[-0.6, 0.6]$                                                                                                      \\
                          & $\omega$ & 0       & $[-0.6, 0.6]$                                           & $[-1.0, 1.0]$                                                                 \\ \midrule
\multirow{5}{*}{\makecell{Behavior\\Commands}}     & $f$        & 2       & \multicolumn{2}{c}{$[1.5, 3.5]$}                                                                                                        \\
                          & $l$     & 0.15    & \multicolumn{2}{c}{$[0.1, 0.35]$}                                                                                                       \\ \cmidrule{2-5}
  & $h$           & 0       & \multicolumn{2}{c}{$[-0.3, 0] $}                                                                                                        \\
                          & $p$            & 0       & \multicolumn{2}{c}{$[0, 0.4]$}                                                                                                          \\
                          & $w$             & 0       & \multicolumn{2}{c}{$[-1.0, 1.0]$}                                                                                                       \\ \bottomrule
\end{tabular}
    % \begin{tabular}{@{}c|cccc@{}} \toprule
    %     Group                              &  Term       & Default & Range \\ \midrule
    %     \multirow{3}{*}{Movement} & linear velocity $v_x$    & 0  & $[-0.6, 0.6] \overset{\text{cur}}{\to} [-0.6, 2.0]$ \\
    %                               & linear velocity $v_y$    & 0     & $[-0.6, 0.6]$ \\
    %                               & angular velocity $\omega$ & 0  & $[-0.6, 0.6] \overset{\text{cur}}{\to} [-1.0, 1.0]$ \\ \midrule
    %     \multirow{2}{*}{Foot}     & gait frequency $f$       & 2               & $[1.5, 3.5]$  \\
    %                               & foot swing height $l$   & 0.15            & $[0.1, 0.35]$ \\ \midrule
    %     \multirow{3}{*}{Posture}  & body height $h$         & 0               & $[-0.3, 0] $  \\
    %                               & body pitch $p$          & 0               & $[0, 0.4]$    \\
    %                               & waist yaw $w$           & 0               & $[-1.0, 1.0]$ \\ \bottomrule
    % \end{tabular}
\end{table}

\subsection{Single Command Tracking}
We first analyze each command separately while keeping all other commands held at their default values. The results are shown in \tb{tab:Single commands}.
It is easily observed that the tracking errors in the walking and standing gaits are significantly lower than those in the jumping and hopping, with hopping exhibiting the largest tracking errors.
For hopping gaits, the robot may fall during the tracking of specific commands, like high-speed tracking, body pitch, and waist-yaw control.
This can be attributed to the fact that hopping requires rather high stability. Moreover, the complex postures and motions further exacerbate the risk of instability. Consequently, the policy prioritizes learning to maintain the balance, which, to some extent, compromises the accuracy of command tracking.

We conclude that the tracking accuracy of each gait aligns with the training difficulty of that gait in simulation. For example, the walking and standing patterns can be learned first during training, while the jumping and hopping gaits appear later and require an extended training period for the robot to acquire proficiency.
Similarly, the tracking accuracy of robots under low velocity is significantly better than those under high velocity, since 1) the locomotion skills under low velocity are much easier to master, and 2) the dynamic stability of the robot decreases at high speeds, leading to a trade-off with tracking accuracy.

We also found that the tracking accuracy for longitudinal velocity commands $v_x$ surpasses that of horizontal velocity commands $v_y$, which is due to the limitation of the hardware configuration of the selected Unitree H1 robots. In addition, the {foot swing height} $l$ is the least accurately tracked.
Furthermore, the tracking reward related to foot placement outperforms the tracking performance associated with posture control, since adjusting posture introduces greater challenges to stability. In response, the policy adopts more conservative actions to mitigate balance-threatening postural changes.
% In contrast, the influence of foot placement on stability is comparatively less pronounced, allowing for more precise tracking.

\begin{table}[t]
\setlength{\abovecaptionskip}{0.cm}
\setlength{\belowcaptionskip}{-0.cm}
\centering
\caption{\small \textbf{Single command tracking error.} The tracking errors for foot commands are calculated over a complete gait cycle, and the remaining ones are over one environmental step. For standing gait, we only tested the body height, body pitch, and waist yaw tracking error. $E^\text{high}$ and $E^\text{low}$ represents high-speed ($v_x > 1m/s$) and low-speed ($v_x \le 1m/s$) modes categorized by the linear velocity $v$. 
The tracking error is computed by sampling each command in a predefined range (\tb{tab:commands}) while keeping all other commands held at their default values.}
\label{tab:Single commands}
\resizebox{\columnwidth}{!}{
\begin{tabular}{@{}c|cccc|cc|ccc@{}}
\toprule
\multirow{3}{*}{Gait} & \multicolumn{4}{c|}{Movement} & \multicolumn{2}{c|}{Foot} & \multicolumn{3}{c}{Posture} \\
\cmidrule(l){2-5} \cmidrule{6-7} \cmidrule{8-10} 
& \multirow{2}{*}{\makecell{$E_{v_x}^\text{low}$\\($m/s$)}} & \multirow{2}{*}{\makecell{$E_{v_x}^\text{high}$\\($m/s$)}} & \multirow{2}{*}{\makecell{$E_{v_y}$\\($m/s$)}} & \multirow{2}{*}{\makecell{$E_{\omega}$\\$rad/s$}} & \multirow{2}{*}{\makecell{$E_{f}$\\($HZ$)}} & \multirow{2}{*}{\makecell{$E_{l}$\\($m$)}} & \multirow{2}{*}{\makecell{$E_{h}$\\($m$)}}  & \multirow{2}{*}{\makecell{$E_{p}$\\($rad$)}} & \multirow{2}{*}{\makecell{$E_{w}$\\($rad$)}}   \\ 
&  &  &  &  &  &  &  &  &    \\ 
\midrule
Standing  & - & - & - & - & - & - & 0.035 & 0.047 & 0.022  \\
Walking   & 0.030 & 0.216 & 0.085 & 0.054 & 0.028 & 0.011 & 0.064 & 0.038 & 0.075  \\
Jumping  & 0.090 & 0.532 & 0.069 & 0.077 & 0.027 & 0.012 & 0.058 & 0.048 & 0.022 \\
Hopping   & 0.033 & - & 0.046 & 0.078 & - & - & 0.103 & - & - \\
\bottomrule
\end{tabular}}
\end{table}

\begin{table*}[t]
\setlength{\abovecaptionskip}{0.cm}
\setlength{\belowcaptionskip}{-0.cm}
\centering
\caption{\small \textbf{Tracking errors with different intervention strategies under the walking gait}. We evaluate three upper-body intervention training strategies: Noise (\our), the AMASS dataset, and no intervention at all. The tracking errors across various task and behavior commands reflect the intervention tolerance, \textit{i.e.}, the ability of precise locomotion control under external intervention.}
\label{tab:Intervetion Tracking Error}
\begin{tabular}{c|c|ccc|cc|ccc}
\toprule
\multirow{3}{*}{Training Strategy} & \multirow{3}{*}{Intervention Task} & \multicolumn{3}{c|}{Task Commands}                        & \multicolumn{5}{c}{Behavior Commands}\\ \cmidrule{3-10}
 & & \multicolumn{3}{c|}{Movement}                        & \multicolumn{2}{c|}{Foot}          & \multicolumn{3}{c}{Posture}                         \\ \cmidrule{3-10}
                                      &                                      &$E_{v_x}$ ($m/s$)     & $E_{v_y}$ ($m/s$)   & $E_{\omega}$ ($rad/s$)    & $E_{f}$ ($Hz$)         & $E_{l}$ ($m$)         & $E_{h}$ ($m$)        & $E_{p}$ ($rad$)     & $E_{w}$ ($rad$)         \\ \midrule
\multirow{3}{*}{\makecell{Noise Curriculum\\(\our)}}        & Noise                        & \textbf{0.0483} & \textbf{0.0962} & \textbf{0.1879} & \textbf{0.0471} & \textbf{0.0542} & \textbf{0.0402} & \textbf{0.0432} & \textbf{0.0552} \\
                                      & AMASS                                & \textbf{0.0391} & \textbf{0.0920} & \textbf{0.1039} & \textbf{0.0464} & \textbf{0.0543} & \textbf{0.0387} & \textbf{0.0364} & \textbf{0.0540} \\
                                      & None                                 & \textbf{0.0264} & \textbf{0.0863} & \textbf{0.0543} & \textbf{0.0447} & \textbf{0.0522} & 0.0372          & 0.0375          & 0.0475          \\ \cmidrule{1-10}
\multirow{3}{*}{AMASS}                & Noise                        & 0.1697          & 0.1055          & 0.2156          & 0.0621          & 0.0542          & 0.0620          & 0.0812          & 0.0694          \\
                                      & AMASS                                & 0.0567          & 0.0965          & 0.1593          & 0.0466          & 0.0555          & 0.0579          & 0.0458          & 0.0554          \\
                                      & None                                 & 0.0645          & 0.0916          & 0.0802          & 0.0460          & 0.0531          & 0.0577          & 0.0455          & 0.0568          \\ \cmidrule{1-10}
\multirow{3}{*}{No Intervention}                 & Noise                        & 0.8658          & 0.7511          & 0.9116          & 0.1930          & 0.1913          & 0.1658          & 0.3622          & 0.2241          \\
                                      & AMASS                                & 0.6299          & 0.4026          & 0.5758          & 0.2245          & 0.2527          & 0.1305          & 0.2367          & 0.1112          \\
                                      & None                                 & 0.0755          & 0.1076          & 0.1151          & 0.0450          & 0.0678          & \textbf{0.0255} & \textbf{0.0211} & \textbf{0.0380} \\ \bottomrule
\end{tabular}
\end{table*}

\begin{table}[t]
\setlength{\abovecaptionskip}{0.cm}
\setlength{\belowcaptionskip}{-0.cm}
\centering
\caption{ \small
\textbf{Averaged foot displacement under intervention}. We compare foot displacement $D_\text{cmd}$ of different training strategies under various intervention tasks, which computes the total movement of both feet in one episode with sampled posture behavior commands.
}
\label{tab:Intervention Mean Foot Movement}
\resizebox{\linewidth}{!}{
\begin{tabular}{ccccc}
\toprule
Training Strategy                 & Intervention Task     & $D_{h}$ ($m/s$)                  & $D_{p}$ ($m/s$)      & $D_{w}$ ($m/s$)       \\ \midrule
\multirow{3}{*}{\makecell{Noise Curriculum\\(\our)}}  & Noise & \textbf{0.0339}             & \textbf{0.0892} & \textbf{0.0199} \\
                       & AMASS         & \textbf{0.0454}             & \textbf{0.0728} & \textbf{0.0196} \\
                       & None          & \textbf{0.0003}             & \textbf{0.0016} & \textbf{0.0007} \\ \midrule
\multirow{3}{*}{AMASS only} & Noise         & 2.0815                      & 2.8978          & 3.2630          \\
                       & AMASS         & 0.0536                      & 0.1743          & 0.0396          \\
                       & None          & 0.0139                      & 0.0160          & 0.0013          \\ \midrule
\multirow{3}{*}{No Intervention}  & Noise         & 17.5358                     & 17.9732         & 25.7132         \\
                       & AMASS         & 25.3802 & 26.3496         & 21.3078         \\
                       & None          & 0.0159  & 1.7065          & 1.7152          \\ \bottomrule
\end{tabular}}
\end{table}

\subsection{Command Combination Analysis}
To provide an in-depth analysis of the command space and to 
reveal the underlying interaction of various commands under different gaits.
Here, we aim to analyze the \emph{orthogonality} of commands based on the interference or conflict between the tracking errors of these commands across their reasonable ranges. For instance, when we say that a set of commands are \emph{orthogonal}, each command does not significantly affect the tracking performance of each other in its range. To this end, we plot the tracking error $E_\text{cmd}$ as heat maps, generated by systematically scanning the command values for each pair of parameters, revealing the correlation of each command.
We leave the full heat maps at \ap{ap:heatmaps}, and conclude our main observation for all gaits.

\noindent\textbf{Walking.} Walking is the most basic gait, which preserves the best performance of the robot hardware.
\begin{itemize}[leftmargin=*]
    \item The {linear velocity} $v_x$, the {angular velocity yaw} $\omega$, the {body height} $h$, and the {waist yaw} $w$ are orthogonal during walking.
    \item When the {linear velocity} $v_x$ exceeds $1.5m/s$, the orthogonality between $v_x$ and other commands decreases due to reduced dynamic stability and the robot's need to maintain body stability over tracking accuracy.
    \item The {gait frequency} $f$ shows discrete orthogonality, with optimal tracking performance at frequencies of 1.5 or 2. High-frequency gait conditions reduce tracking accuracy.
    \item The {linear velocity} $v_y$, the {foot swing height} $l$, and the {body pitch} $p$ are orthogonal to other commands only within a narrow range.
\end{itemize}

\noindent\textbf{Jumping.} The command orthogonality in jumping is similar to walking, but the overall orthogonal range is smaller, due to the increased challenge of the jumping gait, especially in high-speed movement modes.
During each gait cycle, the robot must leap forward significantly to maintain its speed. To execute this complex jumping action continuously, the robot must adopt an optimal posture at the beginning of each cycle. Both legs exert substantial torque to propel the body forward. Upon landing, the robot must quickly readjust its posture to maintain stability and repeat the actions. Consequently, during movement, the robot can only execute other commands within a relatively narrow range.

\noindent\textbf{Hopping.}
The hopping gait introduces more instability, and the robot's control system must focus more on maintaining balance, making it difficult to simultaneously handle complex, multi-dimensional commands.
\begin{itemize}[leftmargin=*]
    \item Hopping gait commands lack clear orthogonal relationships.
    \item Effective tracking is limited to the x-axis {linear velocity} $v_x$, the y-axis {linear velocity} $v_y$, the {angular velocity yaw} $\omega$, and the {body height} $h$.
    \item Adjustments to $h$ can be understood that a lower body height improves dynamic stability, therefore, it plays a positive role in maintaining the target body posture.
    % enhancing the robot's hopping performance.
\end{itemize}

\noindent\textbf{Standing.} As for the standing gait, we tested the tracking errors of commands related to posture. The results showed that the tracking errors were similar to those observed during walking with zero velocity.

\begin{itemize}[leftmargin=*]
    \item The {waist yaw} $w$ command is almost orthogonal to the other two commands.
    \item As the range of commands increases, orthogonality between the {body height} $h$ and the {body pitch} $p$ decreases. This is because the H1 robot has only one degree of freedom at the waist, limiting posture adjustments to the hip pitch joint.
    \item A 0.3 m decrease of the body height relative to the default height reduces the range of motion of the hip pitch joint to almost zero, hindering precise tracking of body pitch.
\end{itemize}

Furthermore, we conclude that {gait frequency} $f$ highly affects the tracking accuracy of \emph{movement} commands when it is excessively high and low; the \emph{posture} commands can significantly impact the tracking errors of other commands, especially when they are near the range limits.
% We categorize the commands into three groups: \emph{movement}, \emph{foot}, and \emph{posture}. 1) The \emph{movement} commands include the linear velocity $v_x, v_y$ and angular velocity $\omega$, forming the foundational locomotion commands, and are considered the most critical aspect of the tasks. 2) The \emph{foot} commands include the {foot swing height} $l$, which is the least accurately tracked; and the {gait frequency} $f$, which can affect the tracking accuracy of \emph{movement} commands when it is excessively high and low. 3) The \emph{posture} commands, which include body height $h$, the body pitch $p$, and waist yaw $w$, determine the desired body posture, and can significantly impact the tracking errors of other commands, especially when the command is challenging. 
For different gaits, the orthogonality range between commands is greatest in the walking gait and smallest in the hopping gait.

\subsection{Ablation on Intervention Training Strategy}
\label{sec:InterventionExp}
% The three policies use the same random seeds and training time.
To validate the effectiveness of the intervention training strategy on the policy robustness when external upper-body intervention is involved, we compare the policies trained with different strategies, including noise curriculum (\our), filtered AMASS data~\citep{he2024omnih2o}, and no intervention. We test the tracking errors under two different intervention tasks, \textit{i.e.}, uniform noise, AAMAS dataset, along with a no-intervention setup. The results under the walking gait are shown in \tb{tab:Intervetion Tracking Error}, and we leave other gaits in \ap{ap:SingleCommandsTracking-REMAIN}. 
It is obvious that the noise curriculum strategy of \our achieved the best performance under almost all test cases, except the posture-related tracking with no intervention. 
In particular, \our showed less of a decrease in tracking accuracy with various interventions, indicating our noise curriculum intervention strategy enables the control policy to handle a large range of arm movements, making it very useful and supportive for loco-manipulation tasks.
In comparison, the policy trained with AMASS data shows a significant decrease in the tracking accuracy when intervening with uniform noise, due to the limited motion in the training data. The policy trained without any intervention only performs well without external upper-body control.

It is worth noting that when intervention training is involved, the tracking error related to the movement and foot is also better than those of the policy trained without intervention, and \our provides the most accurate tracking. This shows that intervention training also contributes to the robustness of the policy. During our real robot experiments, we further observed that the robot behaves with a harder force when in contact with the floor, indicating a possible trade-off between motion regularization and tracking accuracy when involving intervention.

\noindent\textbf{Stability under standing gait.}
Adjusting posture in the standing state introduces additional requirements for stability, since the robot pacing to maintain balance may increase the difficulty of achieving manipulation tasks that require stand still. To investigate the necessity of noise curriculum for manipulation, we further measured the averaged foot displacement (in meters) under the standing gait, which computes the total movement of both feet in one episode (20 seconds) while tracking the posture behavior commands. Results in \tb{tab:Intervention Mean Foot Movement} show that \our exhibits minimal foot displacement. On the contrary, the strategy trained on AMASS data requires frequent small steps to adjust the posture and maintain stability for noise interventions. 
Without intervention training, the policy tends to tip over when involving intervention, leading to failure of the entire task.

%  鲁棒性测试的结果分析
\begin{figure}[t]
    \centering
    \includegraphics[width=\linewidth]{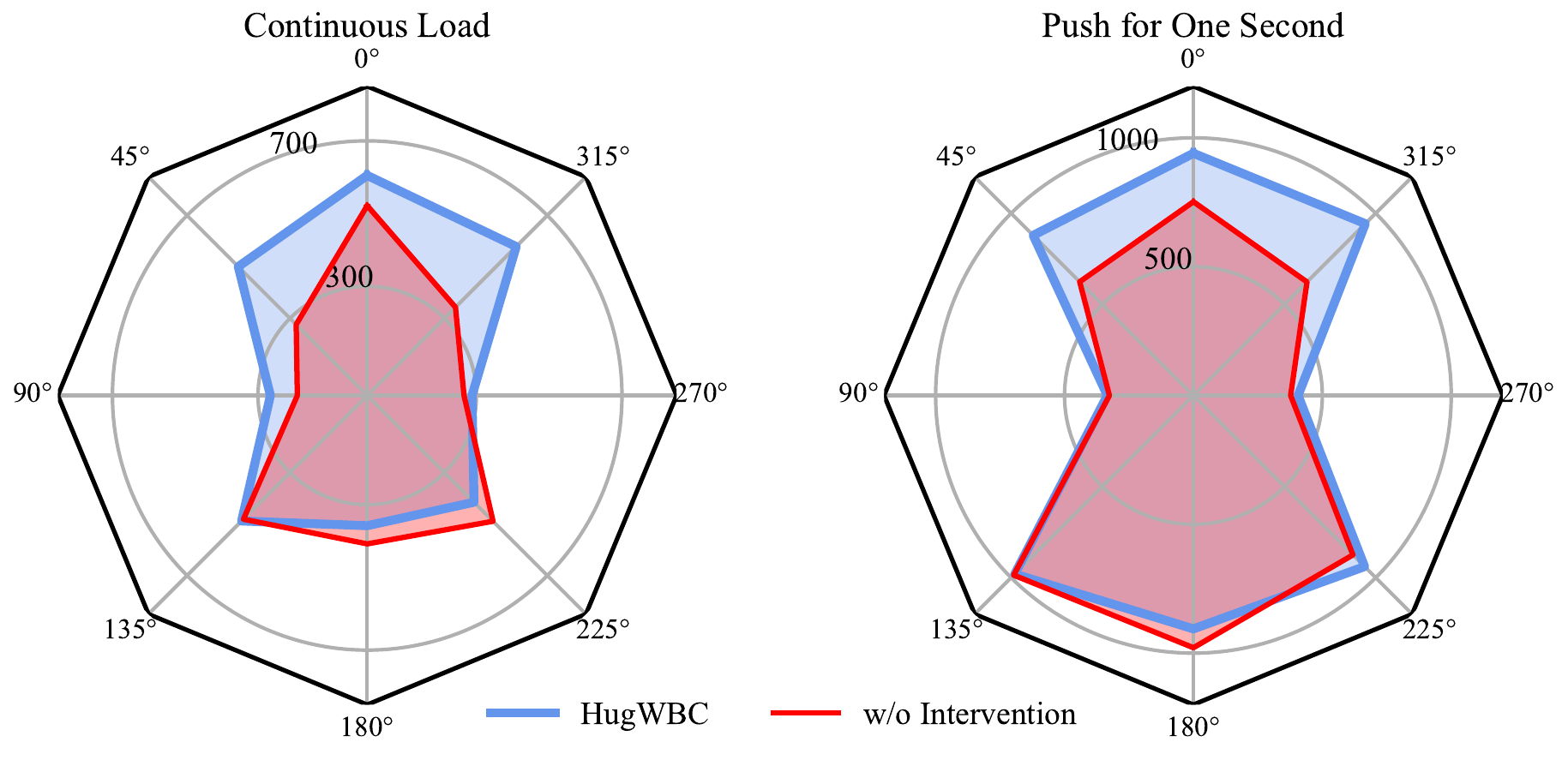}
    \vspace{-13pt}
    \caption{\small \textbf{External disturbance tolerance}. Left: A constant and continuous force is applied to the robot. Right: A one-second force is exerted on the robot. The experiment is conducted under a standing gait with default commands. If the robot's survival ratio exceeds $98\%$, it is deemed capable of tolerating such external disturbance. 
    The survival ratio computes the trajectory ratio of non-termination (ends of timeout) during 4096 rollouts.}
    \label{fig:Robust}
    \vspace{-12pt}
\end{figure}
\noindent\textbf{Robustness for external disturbance.}
Finally, we test the contribution of intervention training and noise curriculum to the robustness of external disturbance. In particular, we evaluated the robot's maximum tolerance to external disturbance forces in eight directions and compared the policy trained without intervention. Results illustrated in \fig{fig:Robust} demonstrate that \our preserves greater tolerance for external disturbances in both pushing and loading scenarios across most of the directions. The reason behind this is that the intervention brings the robot exposed to various disturbances originating from its upper body, and thereby enhances the overall stability by dynamically adjusting leg strength.

% \our has a significantly higher tolerance for external disturbance forces in almost all directions compared to the strategy without intervention training.
% This is attributed to the fact that, during large-scale noise intervention training, the robot effectively explored a wide range of extreme scenarios and learned to enhance body stability by adjusting leg movements.

\subsection{Real-World Experiments}
We deploy \our on a real-world robot to verify its effectiveness. In \fig{fig:teaser}, we illustrate the humanoid capabilities supported by \our, showing the versatile behavior of the Unitree H1 robot. In particular, we demonstrate the intriguing potential of the comprehensive task range that \our is able to achieve, with a flexible combination of commands in high dynamics. To qualitatively analyze the performance of \our, we estimate the tracking error of two pose parameters (body pitch $p$ and waist rotation $w$ from the motor readings) on real robots, since other commands are hard to measure without a highly accurate motion capture system. The results are shown in \tb{tb:track-real}, where $E^{\text{real}}_{\text{cmd}}$ illustrates the tracking error of the posture command.
We observe that the tracking error in real-world experiments is slightly higher than in simulation environments, primarily due to sensor noise and the wear of the robot's hardware. Among different gaits, the tracking error for the waist rotation $w$ is smaller compared to that for the body pitch $p$, as waist control has less impact on the robot’s overall stability. In both error tests, the jumping gait exhibited the smallest $E_{cmd}$, while the walking gait showed slightly higher errors, consistent with the findings observed in the simulation environment.

\begin{table}[t]
\centering
\caption{\small \textbf{Tracking error in real world.} We conducted five tests to measure the tracking error for each command under three gaits. The tracking error for each command was calculated during each control step. The tested commands gradually increased from the minimum to the maximum values within a predefined range, while the remaining commands were kept at their default values.} % To account for the impact of communication delays on the actual tracking error, we introduced a 0.1-second delay in the command execution.
\label{tb:track-real}
\begin{tabular}{c|cc} \toprule
Gait     & $E_p^{\text{real}}$ & $E_w^{\text{real}}$ \\ \midrule
Standing & 0.0712 $\pm$ 0.0425 & 0.0718 $\pm$ 0.0614 \\
Walking  & 0.1006 $\pm$ 0.0581  & 0.0571 $\pm$ 0.0489 \\
Jumping  & 0.0674 $\pm$ 0.0569  & 0.0552 $\pm$ 0.0469 \\ \bottomrule
\end{tabular}
\end{table}

\section{Conclusion and Limitations} 
\label{sec:conclusion} 
We present a unified and general humanoid whole-body controller. Through an extended command space and intervention training, \our enables versatile gait control while supporting external upper-body control, which can serve as a basic controller for extensive loco-manipulation tasks. Future works can adopt \our to control various humanoid robots, or take the policy trained by \our as a unified low-level controller to build a high-level planner to achieve complicated tasks.

\section*{Acknowledgments}
We thank Jingxiao Chen, Xinyao Li, Jiahang Cao, and Xin Liu for their kind support of upper body control, motion generation, and demo recording. We thank 
anonymous reviewers for their kind suggestions. We thank Unitree for their help on the hardware.

%% Use plainnat to work nicely with natbib. 

\bibliographystyle{plainnat}
\bibliography{references}

\clearpage
\newpage
\appendices

\section{Extended Background}
\label{ap:bk}
% \subsection{Humanoid Whole-Body Control as Reinforcement Learning}
% To support various high-level functionalities and allow the humanoid robot to perform complicated tasks, we will need a whole-body controller that serves as a basic component for controlling the robots.
% Formally, we can define the humanoid whole-body control tasks as a Partial Observable Markov Decision Process (POMDP) defined by tuple $\mathcal M = \{\mathcal{S}, \mathcal{O}, \mathcal{C}, \mathcal{A}, T, r, \gamma\}$, where $\mathcal{S}, \mathcal{O}, \mathcal{C}, \mathcal{A}$ are the state, observation, command and action spaces respectively. $T(\cdot|s_t, a_t)$ is the transition density when action $a_t$ received at state $s_t$. Reward functions $r$ is defined typically as the negative distances $D(\cdot)$ or the positive similarity $S(\cdot)$ between the current robot state and the desired robot state $s_t^{c_t}$ which is decided by the actual command $c_t$:
% \begin{equation}
%     \centering
%     \begin{aligned}
%     &r_{\text{negative}}(s_t, a_t, c_t) = - D(s_t, 
%     s_t^{c_t}) \\
%     &r_{\text{positive}}(s_t, a_t, c_t) = S(s_t, s_t^{c_t})
%     \end{aligned}
%     ~.
% \end{equation}
% And $\gamma$ is the discount factor. 

% \minghuan{The two subsections need to be rephrased as I just copied from another doc.}

\subsection{Proximal Policy Optimization}
Proximal policy optimization (PPO)~\citep{schulman2017proximal} is one of the popular algorithms that solve reinforcement learning (RL) problems. 
The goal of RL is to find the optimal policy $\pi^*: \mathcal{O} \times \mathcal{C} \to \mathcal{A}$ for command tracking that maximizes the expected discounted return:
\begin{equation}
    \pi ^{*} = \mathop{\arg \max}_{\pi} \mathbb E_{\pi}\left[\sum_{t=0}^{\infty} \gamma^t r(o_t, a_t, c_t)\right]
\end{equation}
The basic idea behind PPO is to maximize a surrogate objective that constrains the size of the policy update. In particular, PPO optimizes the following objective:
\begin{equation}
    \mathcal L_{\text{policy}} = \bbE_{\pi}[\min(rA, \text{clip}(r,1-\epsilon,1+\epsilon)A)],
\end{equation}
where $r=\frac{\pi(a|o,c)}{\pi_{old}(a|o,c)}$ defines the probability ratio of the current policy and the old policy at the last optimization step, $A$ is the advantage function, which is calculated by learning the value function:
\begin{equation}
\begin{aligned}
    &\mathcal L_{\text{value}} = \bbE_\pi\left[\|V_\pi(o,c) - V^{\text{targ}}(o,c)\|^2\right]~, \\
    &A(o,a,c) = \sum_{t} \gamma^t r(o_t,a_t,c_t) - V(o,c) |_{o_0=o,a_0=a,c_0=c}~,
\end{aligned}
\end{equation}
where $V^{\text{targ}}$ is the target value function, defined as the expected return on the state $o,c$:
\begin{equation}
    V^{\text{targ}}(o,c) = \bbE_{\pi}\left [\sum_{t} \gamma^t r(o_t,a_t,c_t) |o_0=o,c_0=c\right]
\end{equation}

% In our work, we utilize a reinforcement learning algorithm, PPO, and an online imitation learning algorithm, DAgger, to learn the control policies under our framework. In this section, we briefly introduce their basics.
% \subsection{Online Imitation Learning and Teacher-Student Training}
% In general, imitation learning (IL) \cite{ross2011reduction,liu2020energy} studies the task of learning from expert demonstrations. In this work, instead of learning from offline expert data, we refer to an online IL method, dataset aggregation (DAgger)~\cite{ross2011reduction} such that we request an online expert to provide the demonstrated action. Formally, the goal is to train a student policy $\hat{\pi}$ minimizing the action distance between the expert policy $\pi_E$ under its encountered states:
% \begin{equation}
% \hat{\pi} = \arg\min_{\pi\in\Pi}\mathbb E_{s\sim d_{\pi}}[\ell(s,\pi)].
% \label{eq:imitation}
% \end{equation}
% Here, $\ell$ is the mean square error in practice.

\subsection{Asymmetric Training}
The asymmetric training introduces a separate encoder to estimate the key privileged information $s^{\text{key}}$ from $k$-step history proprioceptive observations $h^{k}$, which is trained by an estimation loss $\mathcal{L}_{\text{est}}$: 
\begin{equation}
    \mathcal{L}_{\text{est}} = \bbE_\pi\left[\|\mathcal E_{\pi}(h^k) - s^{\text{key}}\|^2\right]
\label{eq:asym_detail}
\end{equation}

\section{Implementations Details}
% \subsection{Details of Network Architecture}

\subsection{Unitree H1 DOF}
\begin{figure}
    \centering
    \includegraphics[width=0.9\linewidth]{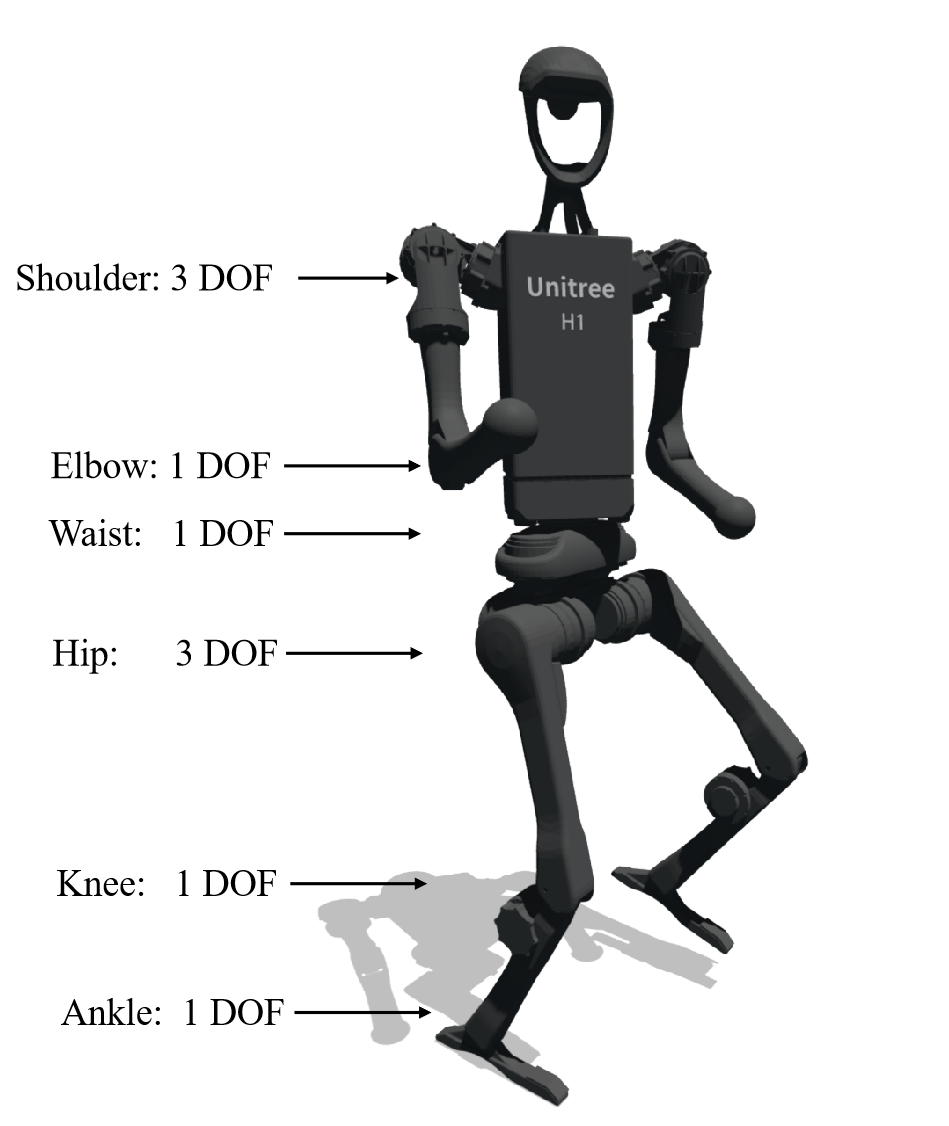}
    \caption{\small DOF demonstration of Unitree H1.}
    \label{fig:H1-dof}
\end{figure}

The Unitree H1, as demonstrated in \fig{fig:H1-dof}, has 19 DoFs in total, including two 3-DOF shoulder joints,
two elbow joints, one waist joint, two 3-DOF hip joints, two
knee joints, and two ankle joints. 

\subsection{Commands Space of The Hopping Gait}
\label{ap:Hopping}
Hopping, characterized by a single foot consistently maintaining ground contact while the other remains in the air, represents an extremely unstable gait that requires coordinated whole-body motor control to maintain balance while tracking task commands. Among the behavior commands, all terms significantly challenge the delicate balance of the robot, except for body height, which poses minimal disruption. 
Therefore, the command space of hopping gait is designed as $\{v_x, v_y, \omega, h\}$, whose remaining behavior command terms for other gaits turn into regular terms.
The command ranges and default value for gait hopping are illustrated in \tb{tab:hopcommands}.
\begin{table}[t]
    \centering
    \caption{\small Ranges and default values of commands for gait hopping.} 
    \begin{tabular}{@{}c|ccc@{}} \toprule
        Group                              &  Term       & Default         &  Range \\ \midrule
        \multirow{3}{*}{Movement} & linear velocity $v_x$    & 0               & $[-0.6, 0.6]$ \\
                                  & linear velocity $v_y$    & 0               & $[-0.6, 0.6]$ \\
                                  & angular velocity $\omega$ & 0               & $[-0.6, 0.6]$ \\ \midrule
        Posture & body height $h$         & 0               & $[-0.3, 0] $  \\
        \bottomrule

    \end{tabular}
    \label{tab:hopcommands}
\end{table}

\subsection{Foot Trajectory Target}
\label{ap:Foot Target}
There are various methods for robot foot trajectory planning, including Bezier trajectory, polynomial trajectory, and so on. Due to the smoothness provided by polynomial trajectory, they are widely used in the swing trajectory planning of quadruped robots~\citep{LocManMPC2021RAL}. Based on this, a polynomial foot trajectory planner integrated with homogeneous variables $\bar{\phi_i}$ is designed in this paper.
% In the z direction, a quintic polynomial trajectory is used for planning. The specific expression of the quintic polynomial is as follows:
% \yufei{The normal quintic curve is a function of time t, but in this work, in order to fit the instruction space, we replaced time t with a uniform phase variable. Do we need to explain this, or do we just write down the equation}
% \begin{equation}\label{eq:foot_traj}
% \begin{aligned}
% l_z(t) = a_5\bar{\phi_i}^5+a_4\bar{\phi_i}^4+a_3\bar{\phi_i}^3+a_2\bar{\phi_i}^2+a_1\bar{\phi_i}+a_0,
% \end{aligned}
% \end{equation}
% where $a_i, i=0, 1, 2, 3, 4, 5$ is the coefficient of polynomial, and $\bar{\phi_i}$ is homogeneous variables. 
In the $z$-axis, the swing trajectory is divided into two segments: from the starting position $p_{s,z}$ to the highest point $l_t$, and from the $l_t$ to the end position $p_{e, z}$. In this study, a piecewise quintic polynomial is used for the foot trajectory planning. For the $p_{s, z}$ and $p_{e, z}$, it is desirable for the foot to make contact with the ground as smoothly as possible. Therefore, both velocity and acceleration are set to zero at these boundary points. The boundary conditions for the piecewise quintic polynomial trajectory are summarized in the \tb{tab:boundary_condition}. 
The coefficients of a polynomial can be calculated:

% \begin{eqnarray}
\begin{equation}
\resizebox{\columnwidth}{!}{
$
\begin{aligned}
\begin{split}
&\left[
\begin{array}{c}
   p_{s, z} \\
   l_t \\
   0 \\
   0 \\
   0 \\
   0
\end{array}
\right]
= 
&\left[
\begin{array}{cccccc}
    ({\bar{\phi}_i^{0.5}})^5 & ({\bar{\phi}_i^{0.5}})^4 & ({\bar{\phi}_i^{0.5}})^3 & ({\bar{\phi}_i^{0.5}})^2 & \bar{\phi}_i^{0.5} & 1 \\
    ({\bar{\phi}_i^{0.75}})^5 & ({\bar{\phi}_i^{0.75}})^4 & ({\bar{\phi}_i^{0.75}})^3 & ({\bar{\phi}_i^{0.75}})^2 & \bar{\phi}_i^{0.75} & 1 \\
    5({\bar{\phi}_i^{0.5}})^4 & 4({\bar{\phi}_i^{0.5}})^3 & 3({\bar{\phi}_i^{0.5}})^2 & 2{\bar{\phi}_i^{0.5}} & 1 & 0 \\
    5({\bar{\phi}_i^{0.75}})^4 & 4({\bar{\phi}_i^{0.75}})^3 & 3({\bar{\phi}_i^{0.75}})^2 & 2{\bar{\phi}_i^{0.75}} & 1 & 0 \\
    20({\bar{\phi}_i^{0.5}})^3 & 12({\bar{\phi}_i^{0.5}})^2 & 6{\bar{\phi}_i^{0.5}} & 2 &0 & 0 \\
    20({\bar{\phi}_i^{0.75}})^3 & 12({\bar{\phi}_i^{0.75}})^2 & 6{\bar{\phi}_i^{0.75}} & 2 &0 & 0
\end{array}
\right] 
\left[
\begin{array}{c}
    a_{5}^1 \\ 
    a_{4}^1 \\
    a_{3}^1 \\
    a_{2}^1 \\
    a_{1}^1 \\
    a_{0}^1
\end{array}
\right]~,
\end{split}
\\
\begin{split}
&\left[
\begin{array}{c}
   l_t \\
   p_{e, z} \\
   0 \\
   0 \\
   0 \\
   0
\end{array}
\right]
= 
&\left[
\begin{array}{cccccc}
    ({\bar{\phi}_i^{0.75}})^5 & ({\bar{\phi}_i^{0.75}})^4 & ({\bar{\phi}_i^{0.75}})^3 & ({\bar{\phi}_i^{0.75}})^2 & \bar{\phi}_i^{0.75} & 1 \\
    ({\bar{\phi}_i^{1.0}})^5 & ({\bar{\phi}_i^{1.0}})^4 & ({\bar{\phi}_i^{1.0}})^3 & ({\bar{\phi}_i^{1.0}})^2 & \bar{\phi}_i^{1.0} & 1 \\
    5({\bar{\phi}_i^{0.75}})^4 & 4({\bar{\phi}_i^{0.75}})^3 & 3({\bar{\phi}_i^{0.75}})^2 & 2{\bar{\phi}_i^{0.75}} & 1 & 0 \\
    5({\bar{\phi}_i^{1.0}})^4 & 4({\bar{\phi}_i^{1.0}})^3 & 3({\bar{\phi}_i^{1.0}})^2 & 2{\bar{\phi}_i^{1.0}} & 1 & 0 \\
    20({\bar{\phi}_i^{0.75}})^3 & 12({\bar{\phi}_i^{0.75}})^2 & 6{\bar{\phi}_i^{0.75}} & 2 &0 & 0 \\
    20({\bar{\phi}_i^{1.0}})^3 & 12({\bar{\phi}_i^{1.0}})^2 & 6{\bar{\phi}_i^{1.0}} & 2 &0 & 0
\end{array}
\right] 
\left[
\begin{array}{c}
    a_{5}^2 \\ 
    a_{4}^2 \\
    a_{3}^2 \\
    a_{2}^2 \\
    a_{1}^2 \\
    a_{0}^2
\end{array}
\right]~.
\end{split}
\label{eq:polynomial coefficients}
\end{aligned}
$
}
\end{equation}
% \end{eqnarray}

where $p_{s,z}$ is the $z$-coordinate of the start position, $p_{e,z}$ is the $z$-coordinate of the start position and $l_t$ is swing highest position. The piecewise quintic polynomial trajectory $l_t^{\text{target},i}$ is formulated as:
\begin{equation}
\resizebox{\columnwidth}{!}{
$
\begin{aligned}
l_t^{\text{target},i} = \left\{\begin{aligned}
&\frac{6(l_t-p_{s,z})}{(\bar{\phi}_i^{0.75}-\bar{\phi}_i^{0.5})^5}(\bar{\phi_i}-0.5)^5 + \frac{15(p_{s,z} - l_t)}{(\bar{\phi}_i^{0.75}-\bar{\phi}_i^{0.5})^4}(\bar{\phi_i}-0.5)^4 + \frac{10(p_{s,z}-l_t)}{(\bar{\phi}_i^{0.75}-\bar{\phi}_i^{0.5})^3}(\bar{\phi_i}-0.5)^3 + p_{s,z},
&0.5 < \phi_i < 0.75 \\ 
&\frac{6(p_{e, z} - l_t)}{(\bar{\phi}_i^{1.0}-\bar{\phi}_i^{0.75})^5}(1-\bar{\phi_i})^5 + \frac{15(l_t - p_{e, z})}{(\bar{\phi}_i^{1.0}-\bar{\phi}_i^{0.75})^4}(1-\bar{\phi_i})^4 + \frac{10(l_t - p_{e,z})}{(\bar{\phi}_i^{1,0} -\bar{\phi}_i^{0.75})^3}(1-\bar{\phi_i})^3 + l_t ,   & 0.75 < \phi_i < 1.0
\end{aligned}\right.~,
\end{aligned}
$
}
\end{equation}

\begin{table}[t]
\setlength{\abovecaptionskip}{0.cm}
\setlength{\belowcaptionskip}{-0.cm}
    \centering
    \caption{\small \textbf{Boundary conditions} for $z$-direction quintic polynomial trajectory.}
    \label{tab:boundary_condition}
\begin{tabular}{cccc}
\toprule
Time                & Position  & Velocity & Acceleration \\ \midrule
t=$\bar{\phi_i^{0.5}}$  & $p_{s,z}$ & 0        & 0            \\
t=$\bar{\phi_i^{0.75}}$ & $l_t$     & 0        & 0            \\
t=$\bar{\phi_i^{1.0}}$  & $p_{e,z}$ & 0        & 0            \\ \bottomrule
\end{tabular}
\end{table}

\subsection{Details of Intervention Baseline}
In experiment section \ref{sec:InterventionExp}, we compare \our with a baseline policy that is trained with intervention actions sampled from the AAMAS motion dataset.

\noindent\textbf{Motion intervention interpolation.}
Since the frequency of the motion data is different from the control frequency, we interpolate the intervened actions from motion capture datasets to match the control frequency. Formally, at time step $t$, the intervention action is a linear interpolation of the closest two frames from the dataset:
\begin{equation}
    a^{\text{interv}}_{t, \text{dataset}} = (1-\gamma) a_k^{\text{traj}_j} + \gamma a_{k+1}^{\text{traj}_j}
\end{equation}
where 
$$
    \gamma = \frac{f^{\text{traj}_j}\cdot t-T_k^{\text{traj}_j}}{T_{k+1}^{\text{traj}_j}-T_{k}^{\text{traj}_j}}
$$
is the interpolation coefficient,
$T_k^{\text{traj}_j}$ is the original time stamp of the $k$-th frame of $j$-th trajectory in the dataset and $f^{\text{traj}_j}$ is the frequency of the $j$-th trajectory.
The training process keeps the same curricula as described in \eq{eq:noisecurr} by replacing $a^{\text{interv}}_{\text{noise}}$ with $a^{\text{interv}}_{\text{dataset}}$.
% The update time step interval $t_{\text{interval}}$ is $2$, due to the high frequency of frames in the trajectory.

\subsection{Details of Network Architecture}
We deployed an asymmetric training framework. 
\our actor network consists of three key components: a historical state encoder, a state estimator, and a low-level network. The historical state encoder takes in five frames of historical proprioceptive observations $o_t^{\text{his}}$ and outputs an encoded historical vector $z_t$. The state estimator leverages this encoded vector to implicitly estimate linear velocity $\hat{v_t}$, foot clearance $\hat{l_t}$, and body height $\hat{h_t}$ that are often challenging to measure accurately with onboard sensors. Finally, the low-level network processes the $z_t$, the estimated states $\hat{v_t}$, $\hat{l_t}$, $\hat{h_t}$ ,current proprioceptive observations $o_t^{pro}$, the commands $c_t$ and binary indicator $I(t)$, ultimately generating the joint actions $a_t$. A more detailed description of network architecture is shown in the Tab. \ref{tab:network architecture}.

\begin{table}[t]
\setlength{\abovecaptionskip}{0.cm}
\setlength{\belowcaptionskip}{-0.cm}
    \centering
    \caption{\small \textbf{Network architectures.}} 
    \label{tab:network architecture}
\begin{tabular}{llll}
\hline
Module                   & Inputs                                                                                                  & Hidden Layers       & Outputs                           \\ \hline
Historical State Encoder & $o_t^{\text{his}}$                                                                                             & {[}256, 128{]}      & $z_t$                             \\
State Estimator          & $z_t$                                                                                                   & {[}64, 32{]}        & $\hat{v_t}, \hat{l_t}, \hat{h_t}$ \\
Low-Level Network        & \makecell[l]{$z_t, \hat{v_t}, \hat{l_t}, \hat{h_t}$, \\ $o_t^{\text{pro}}, c_t, I(t)$} & {[}256, 128, 64{]}  & $a_t$                             \\
Critic                  & $o_t^{\text{pro}},o_t^{\text{pri}},o_t^{\text{ter}}$                                                                         & {[}512, 256, 128{]} & $V_t$                             \\ \hline
\end{tabular}
\end{table}

\subsection{Policy Learning Time}
The overall policy learning time was 16 hours of wall-clock time, using a single NVIDIA RTX 4090 GPU.

\section{Extended Experiment}
\subsection{Extensive Analysis of Commands Combination}
\label{ap:heatmaps}
We draw heatmaps and line charts to illustrate the tracking accuracy when combining two different commands across their ranges under different gaits, shown in \fig{fig:Heatmaps}.

\begin{table*}[t]
\setlength{\abovecaptionskip}{0.cm}
\setlength{\belowcaptionskip}{-0.cm}
\centering.
\caption{\small \textbf{Tracking error with different intervention strategies under the standing gait and the jumping gait}. We evaluate three upper-body intervention
training strategies: noise curriculum (HUGWBC), the AMASS dataset, and no intervention at all. The tracking errors across various tasks and behavior commands reflect the intervention tolerance, \textit{i.e.}, the ability of precise locomotion control under external intervention.}
\label{tab:Intervention Tracking Error stand jump}
\resizebox{\linewidth}{!}{
\begin{tabular}{c|c|c|ccc|cc|ccc}
\toprule
\multirow{2}{*}{Gait}     & \multirow{2}{*}{Training Strategy}                                                                                                  &   \multirow{2}{*}{Intervention Task}               & \multicolumn{3}{c|}{Movement}                        & \multicolumn{2}{c|}{Foot}          & \multicolumn{3}{c|}{Posture}                           \\ \cmidrule{4-11}
                          &                                                                                                         &                                     &$E_{v_x}$ ($m/s$)     & $E_{v_y}$ ($m/s$)   & $E_{\omega}$ ($rad/s$)    & $E_{f}$ ($Hz$)         & $E_{l}$ ($m$)         & $E_{h}$ ($m$)        & $E_{p}$ ($rad$)     & $E_{w}$ ($rad$)      \\ \midrule
\multirow{9}{*}{Standing} & \multirow{3}{*}{\makecell{Noise Curriculum\\(\our)}}                                                                         & Noise                       & -               & -               & -               & -               & -               & \textbf{0.0291}         & \textbf{0.0662}  & \textbf{0.0605}  \\
                          &                                                                                                         & AMASS                       & -               & -               & -               & -               & -               & 0.0237          & \textbf{0.0611}  & \textbf{0.0624}  \\
                          &                                                                                                         & \multicolumn{1}{c|}{None} & -               & -               & -               & -               & -               & 0.0228          & 0.0476           & 0.0564           \\ \cmidrule{2-11}
                          & \multirow{3}{*}{AMASS}                                                                         & Noise                       & -               & -               & -               & -               & -               & 0.0301 & 0.1222           & 0.0875           \\
                          &                                                                                                         & AMASS                       & -               & -               & -               & -               & -               & \textbf{0.0204} & 0.0782           & 0.0707           \\
                          &                                                                                                         & \multicolumn{1}{c}{None} & -               & -               & -               & -               & -               & \textbf{0.0198}         & 0.0778           & 0.0727           \\ \cmidrule{2-11}
                          & \multicolumn{1}{c}{\multirow{3}{*}{None}} & Noise                       & -               & -               & -               & -               & -               & 0.1931          & 0.4051           & 0.2283           \\
                          & \multicolumn{1}{c}{}                                                                                    & AMASS                       & -               & -               & -               & -               & -               & 0.1357          & 0.2571           & 0.1243           \\
                          & \multicolumn{1}{c}{}                                                                                    & \multicolumn{1}{c}{None} & -               & -               & -               & -               & -               & 0.0213 & \textbf{0.0218} & \textbf{0.0514} \\ \midrule

\multirow{9}{*}{Jumping}  
& \multirow{3}{*}{\makecell{Noise Curriculum\\(\our)}}        & Noise                        & \textbf{0.0886} & \textbf{0.1078} & \textbf{0.1785}   & \textbf{0.0580}  & \textbf{0.0457} & \textbf{0.0411}      & \textbf{0.0471}      & \textbf{0.0527}      \\
                          &                                       & AMASS                                & \textbf{0.0750} & \textbf{0.0729} & \textbf{0.1010}   & \textbf{0.0487}  & \textbf{0.0458} & \textbf{0.0402}      & \textbf{0.0417}      & \textbf{0.0519}      \\
                          &                                       & None                                 & \textbf{0.0504} & \textbf{0.0606} & \textbf{0.0778}   & \textbf{0.0556}  & \textbf{0.0445} & \textbf{0.0417}              & 0.0491               & 0.0511              \\ \cmidrule{2-11} 
                          & \multirow{3}{*}{AMASS}                & Noise                        & 0.3026          & 0.1179          & 0.2736            & 0.0591           & 0.2560          & 0.0424               & 0.1217               & 0.0757               \\
                          &                                       & AMASS                                & 0.0826          & 0.0759          & 0.1104            & 0.0553           & 0.0469          & 0.0428               & 0.0486               & 0.0536               \\
                          &                                       & None                                 & 0.0854          & 0.0743          & 0.0804            & 0.0583           & 0.0461          & 0.0431               & 0.0476               & 0.0551               \\ \cmidrule{2-11} 
                          & \multirow{3}{*}{None}                 & Noise                        & 0.8082          & 0.5533          & 0.8340            & 0.0717           & 0.6358          & 0.1931               & 0.4051               & 0.2283               \\
                          &                                       & AMASS                                & 0.8632          & 0.4105          & 0.6888            & 0.0787           & 0.7720          & 0.1357               & 0.2591               & 0.1243               \\
                          &                                       & None                                 & 0.0711          & 0.0976          & 0.1127            & 0.0625           & 0.3255          & 0.0449      & \textbf{0.0368}      & \textbf{0.0375}      \\ \bottomrule

\end{tabular}}
\end{table*}
\subsection{Commands Tracking with Interventions}
\label{ap:SingleCommandsTracking-REMAIN}
We further show the single command tracking evaluation results for the standing gait and the jumping gait, in \tb{tab:Intervention Tracking Error stand jump}.
On these gaits, \our also achieves the best tracking performance under almost all test cases, except the body pitch and waist yaw tracking with no intervention. In contrast, the policy trained with AMASS data is still limited to handling actions within the scope of that data, and the policy trained without intervention fails with any external upper-body control. We thus can conclude that intervention training applies to a variety of gaits. 

In particular, under the jumping gait, the intervention tasks had a significant impact on the robot tracking performance. This is mainly because jumping gait is more challenging for humanoid robots, which rely heavily on arm swings to complete the motion task. Therefore, when the arm movement is restricted, the robot's performance is notably compromised.
Under the standing gait, \our shows significantly lower posture-related tracking errors compared to the walking and jumping gaits. 

Since hopping is a highly unstable gait, which is rarely used for loco-manipulation tasks and is implemented with an independent policy, we did not involve intervention training for the hopping gait.

\begin{figure*}[htbp]
    \centering
\begin{subfigure}[htbp]{\textwidth}
    \includegraphics[width=\linewidth]{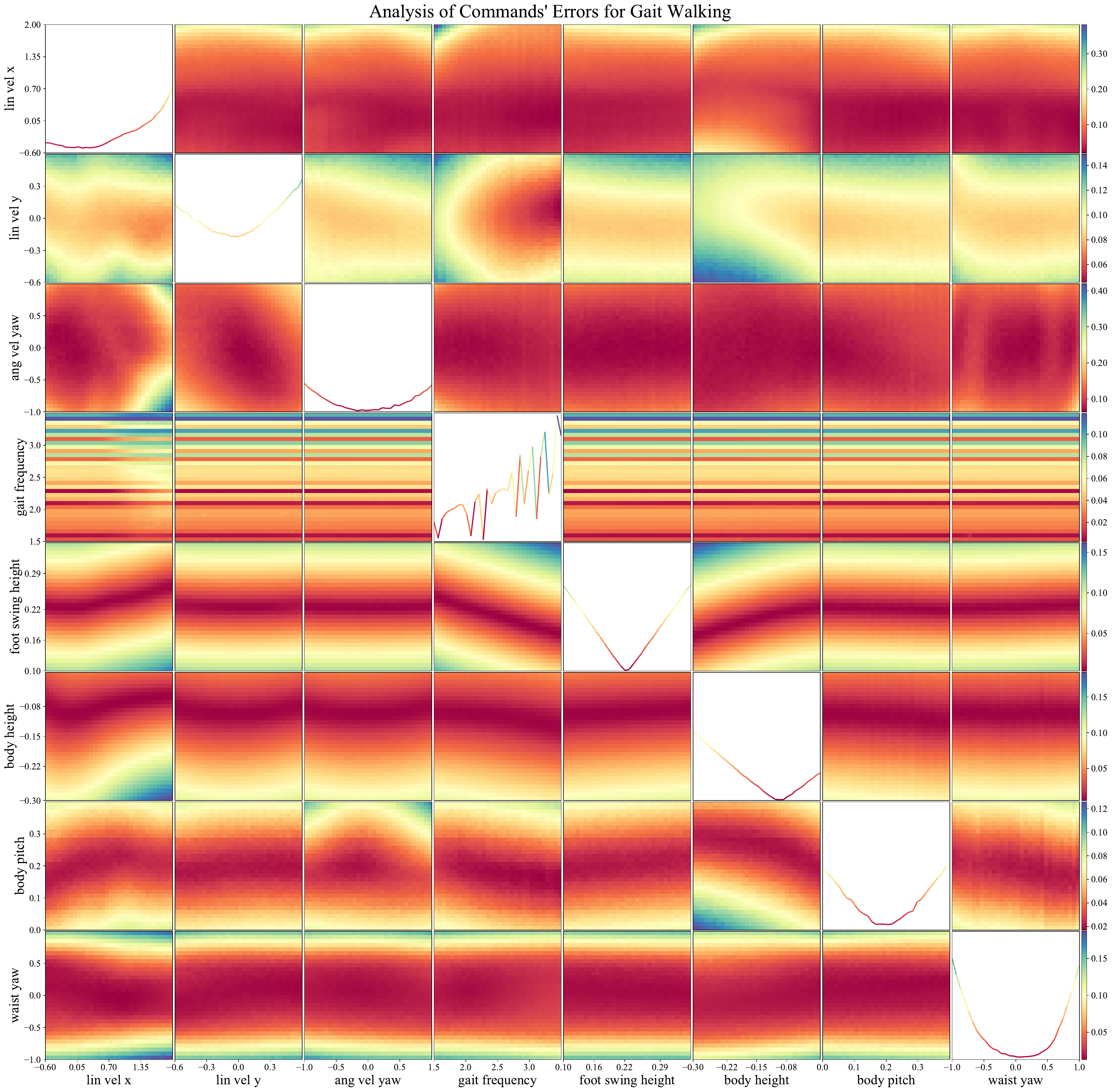}
    \subcaption{Walking.}
    \label{fig:heatmapwalk}
\end{subfigure}
\end{figure*}
\begin{figure*}[htbp]\ContinuedFloat
\begin{subfigure}[htbp]{\textwidth}
    \includegraphics[width=\linewidth]{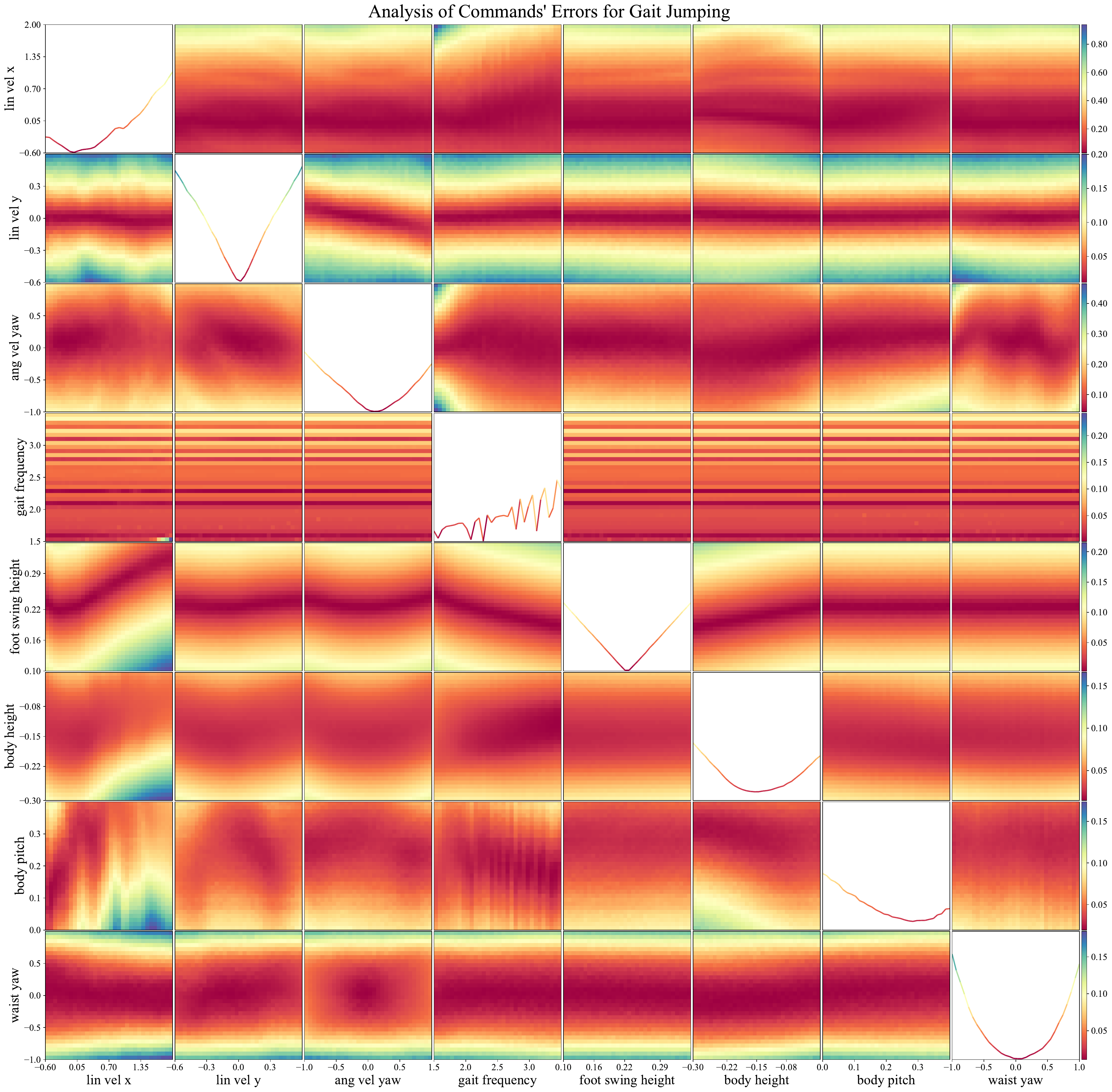}
    \subcaption{Jumping.}
    \label{fig:heatmapjump}
\end{subfigure}
\end{figure*}
\begin{figure*}[htbp]\ContinuedFloat
\begin{subfigure}[htbp]{0.49\textwidth}
    \includegraphics[width=\linewidth]{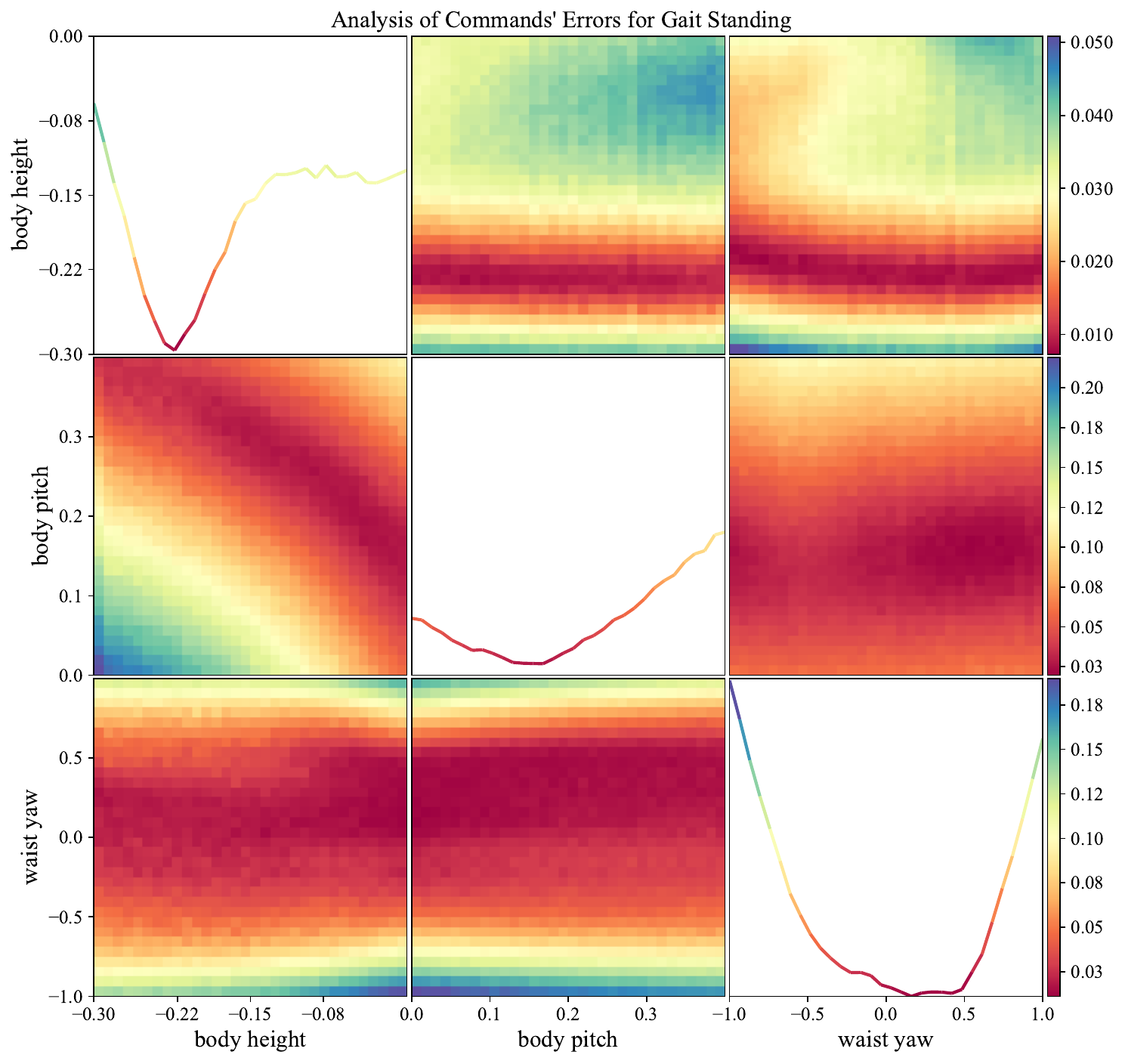}
    \subcaption{Standing.}
    \label{fig:heatmapstand}
\end{subfigure}
\begin{subfigure}[htbp]{0.49\textwidth}
    \includegraphics[width=\linewidth]{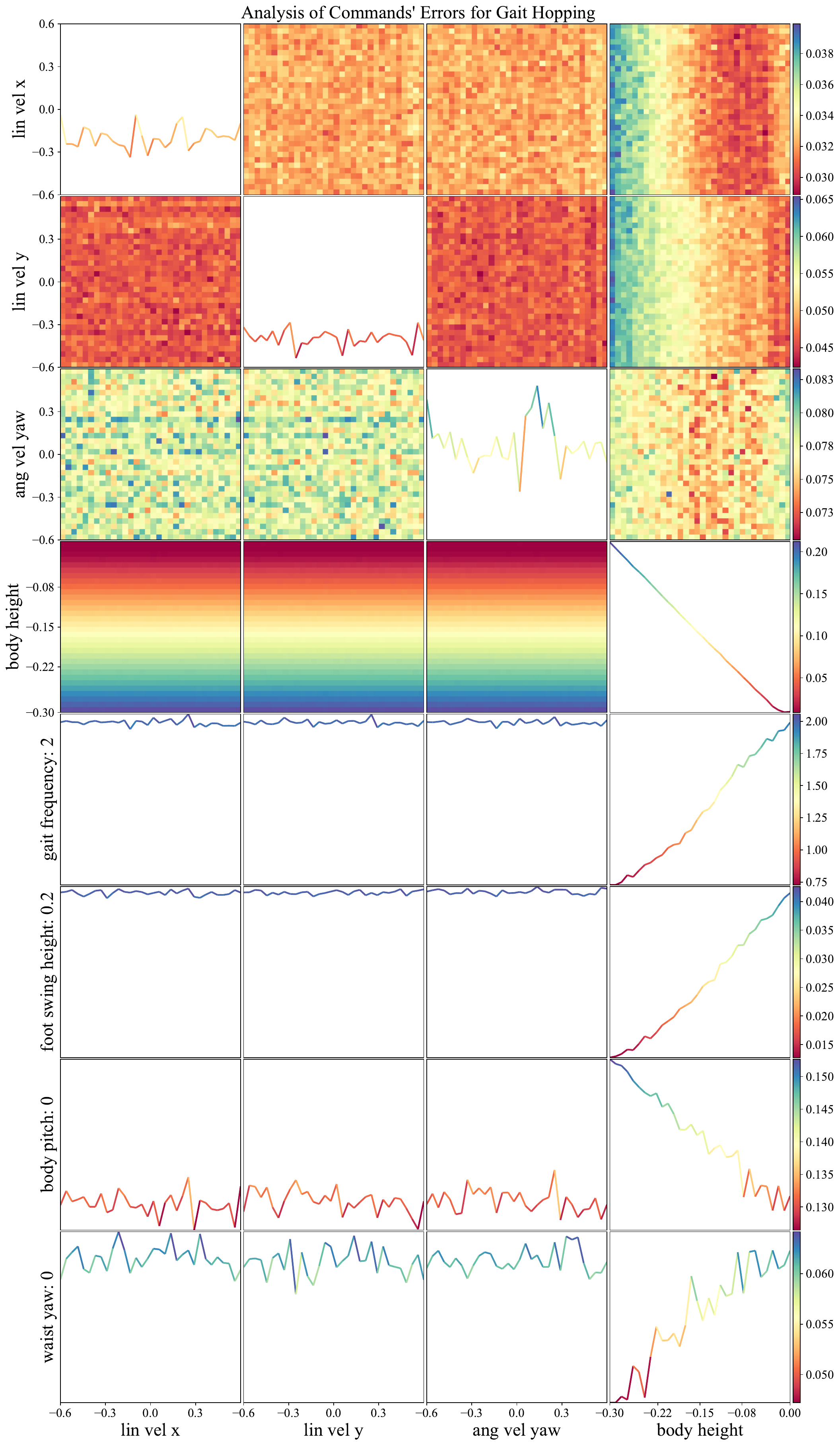}
    \subcaption{Hopping.}
    \label{fig:heatmaphopping}
\end{subfigure}
\caption{\small \textbf{Tracking-error heat maps of command combination under different gaits}. Each column represents one of the following command parameters: \emph{linear velocity x}, \emph{linear velocity y}, \emph{angular velocity yaw}, \emph{gait frequency}, \emph{foot swing height}, \emph{body height}, \emph{body pitch}, and \emph{waist roll}. The standing includes the three series commands for the \emph{body height}, \emph{body pitch}, and \emph{waist yaw}.
For the off-diagonal sub-figures, the range for each command is indicated along the vertical axis (left) and horizontal axis (bottom). The corresponding error values are indicated by ticks on the right-side color bar. 
The colder %darker
the color of the pixel, the larger the tracking error the commands faces, and the color bars in different rows have different ranges of error.}
\label{fig:Heatmaps}
\end{figure*}

\subsection{Comparison with Other Whole-body Controllers}
We compare \our with two \textit{open-sourced SOTA learning-based humanoid whole-body controllers}, HOVER~\citep{he2024hover} and Exbody~\citep{cheng2024expressive} in simulation, shown in \tb{tab:baseline}. Nevertheless, the training and control modes of these controllers rely heavily on motion datasets. 
For example, the command space of ExBody includes target expression goal (upper body) and root movement goal (lower body), which are sampled from trajectories.
% Its training relies heavily on the motion dataset, with command combinations sampled directly from the data. 
Although HOVER features a multi-mode command space, it requires high consistency between different command terms due to the motion tracking task setting.
% ExBody~\citep{cheng2024expressive} also requires motion reference, whose command spaces include body keypoints and joint positions (upper body), and root movement (lower body) which can be decoupled to our posture and movement commands.
To compare them to ours, we keep the upper body of humanoids remaining at the default joint positions as the required reference, and compute the tracking error as in \tb{tab:Single commands}.
Note that we evaluate the performance of \our under the \textit{walking} gait, as the baselines do not support gait switch without reference motion, and \our does not require upper-body reference motion and controls the whole-body joints.
The comparison experiment forces HOVER and ExBody policies to perform tasks beyond their intended design, resulting in poorer performance than demonstrated in their respective papers.

\begin{table}[htbp]
    \centering
    \caption{Single command tracking error comparison with learning based baselines.}
    % under the walking gait, settings are the same as in Table III.}
    % \vspace{-5pt}
    \resizebox{\columnwidth}{!}{
    \begin{tabular}{c|cccc|ccc} \toprule
      Methods & $E_{v_x}^{\text{low}}$ & $E_{v_x}^{\text{high}}$ & $E_{v_y}$ & $E_{\omega}$ & $E_h$ & $E_p$ & $E_w$ \\ \midrule
      HOVER~\citep{he2024hover}   &  0.559 & 1.324 & 0.328 & 0.436 & 0.270 & 0.127 & 0.082\\
      ExBody~\citep{cheng2024expressive}   &  0.109 & 0.242 & 0.114 & 0.587 & 0.145 & 0.122 & 0.097\\
      \textsc{HugWBC} (Ours) & \textbf{0.030} & \textbf{0.216} & \textbf{0.085} & \textbf{0.054} & \textbf{0.064} & \textbf{0.038} & \textbf{0.075} \\ \bottomrule
    \end{tabular}
    }
    \label{tab:baseline}
\end{table}

\end{document}